\documentclass{article}

\usepackage{microtype}
\usepackage{graphicx}
\usepackage{subcaption}
\usepackage{booktabs} 

\usepackage{rotating}

\usepackage{hyperref}

\usepackage[accepted]{icml2026}

\usepackage{amsmath}
\usepackage{amssymb}
\usepackage{mathtools}
\usepackage{amsthm}

\usepackage{multirow}

\usepackage[capitalize,noabbrev]{cleveref}

\theoremstyle{plain}

\theoremstyle{definition}

\theoremstyle{remark}

\usepackage[textsize=tiny]{todonotes}

\icmltitlerunning{Advancing Ligand-based Virtual Screening and Molecular Generation with Pretrained Molecular Embedding Distance}

\begin{document}

\twocolumn[
  \icmltitle{Advancing Ligand-based Virtual Screening and Molecular Generation \\
  with Pretrained Molecular Embedding Distance}



  \icmlsetsymbol{equal}{*}

\begin{icmlauthorlist}
\icmlauthor{Shiyun Wa}{biogen}
\icmlauthor{Yifei Wang}{biogen}
\icmlauthor{Simone Sciabola}{biogen}
\icmlauthor{Ye Wang}{biogen}
\end{icmlauthorlist}

\icmlaffiliation{biogen}{Biogen, Cambridge, MA, USA} 

\icmlcorrespondingauthor{Ye Wang}{ye.wang@biogen.com}

  \icmlkeywords{Machine Learning, ICML}

  \vskip 0.3in
]



\printAffiliationsAndNotice{}  

\begin{abstract}
  Molecular similarity plays a central role in ligand-based drug discovery, such as virtual screening, analog searching, and goal-directed molecular generation. However, traditional similarity measures, ranging from fingerprint-based Tanimoto coefficients to 3D shape overlays, are often computationally expensive at scale or rely on hand-crafted molecular descriptors. Meanwhile, many deep learning approaches to similarity-aware design still depend on similarity-specific supervision or costly data curation, limiting their generality across targets. 
  In this work, we propose pretrained embedding distance (PED) as an effective alternative, computed directly from pretrained molecular models without task-specific training.
  Experimental results show that PED exhibits distinct correlations with traditional similarity metrics, and performs effectively in both ranking molecules for virtual screening and guiding molecular generation via reward design.
  These findings suggest that pretrained molecular embeddings capture rich structural information and can serve as a promising and scalable similarity measurement for modern AI-aided drug discovery.
\end{abstract}

\section{Introduction}
Molecular similarity occupies a central position in ligand-based drug discovery. This approach assumes that molecules with similar structural and pharmacophoric arrangements are likely to occupy the same binding site and elicit similar biological effects \cite{hendrickson1991concepts, maggiora2014molecular}. This concept serves as the critical computational engine for a broad class of modern drug discovery workflows such as virtual screening and generative molecular design. Virtual screening serves as a high-throughput retrieval task, identifying promising binders against the therapeutic target of interest from expansive chemical libraries by ranking candidates against known active templates \cite{KLEBE2006580, gao2023drugclip}. Generative design focuses on the \textit{de novo} construction of molecules that satisfy complex multi-objective property and structure constraints \cite{Haddad2025,edwards2026mclm}. Despite their different formulations, both tasks are fundamentally governed by the evaluation of molecular similarity. In ligand-based virtual screening, similarity serves as the primary ranking heuristic; in reinforcement learning-based molecular generation, the similarity-derived rewards guide the exploration toward bioactive regions of chemical space. Consequently, the development of an accurate, computationally efficient, and general-purpose similarity function remains a vital research direction in modern drug discovery as it enables the navigation of chemical space at an unprecedented scale.

 Traditional measurement requires an appropriate selection of molecular descriptors, similarity coefficients, and weighting schemes \cite{maldonado2006molecular}. These descriptors serve as the mathematical representation of chemical entities, translating constitutional, configurational, and conformational features into a format suitable for computational analysis \cite{nikolova2003approaches}. They span a hierarchy of complexity ranging from 1D physicochemical properties \cite{raevsky2004physicochemical} and 2D topological fingerprints \cite{ECFP}, to sophisticated 3D shape and color\footnote{Chemical color represents the spatial arrangement of pharmacophore features like hydrogen bond donors and acceptors.} alignment such as ROCS \cite{rocs} and ROSHAMBO2 \cite{roshambo2}. However, these approaches face a fundamental trade-off between speed and accuracy. Handcrafted descriptors and 2D fingerprints are often rigid and low-dimensional to capture complex biological mechanisms. 3D approaches, while considered the industry gold standard, require computationally intensive conformer generation and spatial alignment procedures that hinder their application to large-scale molecular libraries.

\begin{figure*}[t]
    \centering
    \includegraphics[width=0.9\linewidth]{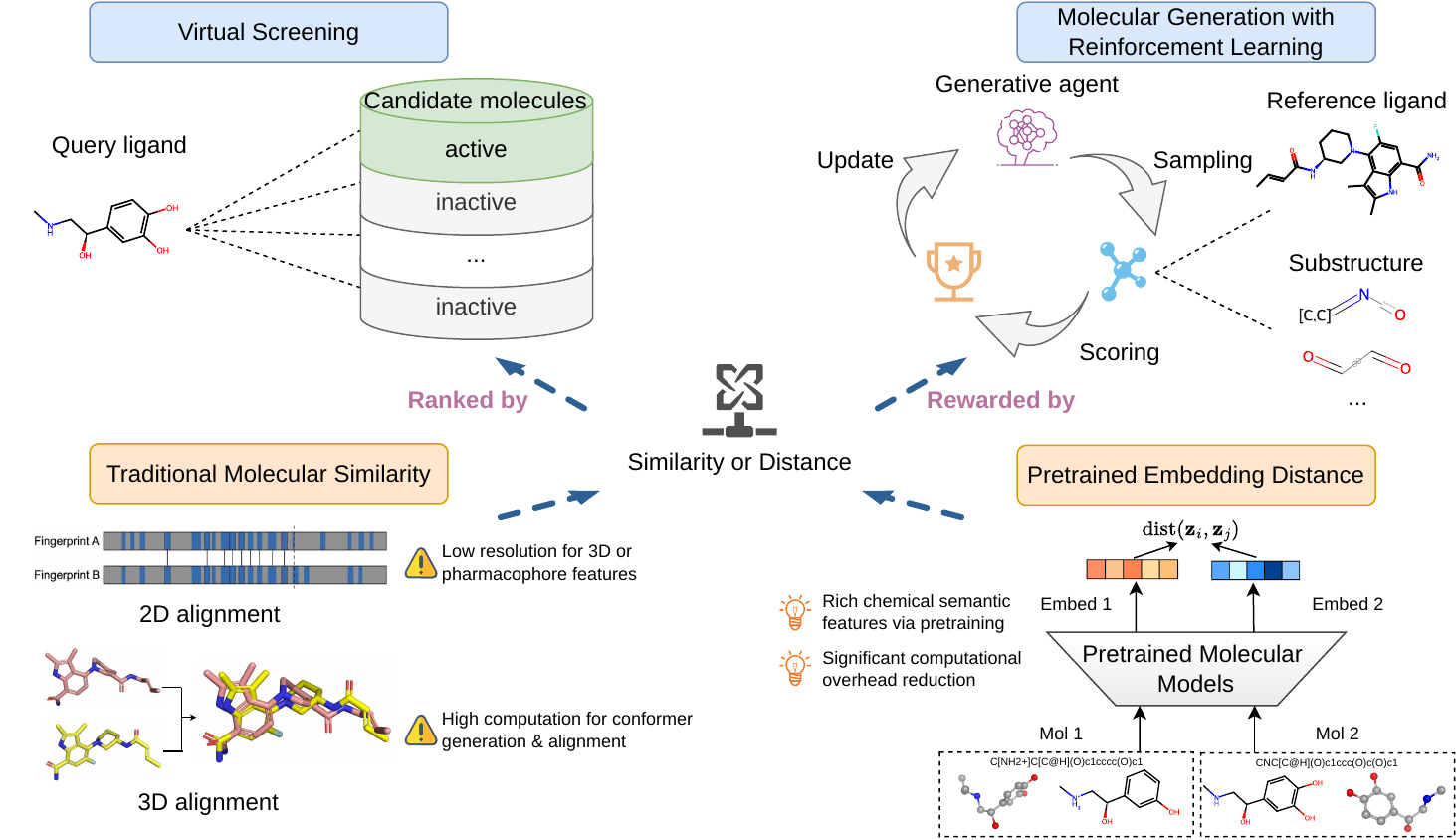}
    \caption{\textbf{PED-Driven Conceptual Framework for Scalable Ligand-based Drug Discovery.} Modern ligand-based drug discovery relies on robust similarity or distance signals to function as scoring functions for prioritizing compounds in virtual screening and as reward for guiding reinforcement learning-based molecular generation. Traditional 2D/3D alignment methods often face inherent representation limitation or prohibitive computational bottlenecks. Instead, this work proposes utilizing Pretrained Embedding Distance (PED) as a highly scalable and computationally efficient alternative. By leveraging embeddings from pretrained models like GeoDiff or MoLFormer, PED captures advanced chemical semantics without explicit alignment, enabling high-resolution discovery at significantly reduced cost.}
    \label{fig:graphic_abstract}
\end{figure*}
 
 Recently, machine learning or deep learning approaches have offered a more flexible and efficient definition of molecular similarity, shifting from manual feature engineering to automated representation learning. To achieve this, many methods utilize supervised or semi-supervised frameworks to learn task-specific similarity functions from experimental data. As a general bottleneck of data-driven methods, these models often fail to generalize well due to high cost and scarcity of high-quality molecular annotations. For example, supervised approaches remain constrained by small, target-specific datasets \cite{McNutt2025, Shah2025}, while semi-supervised or unsupervised approaches such as contrastive learning \cite{atsango2022a, gao2023drugclip, zhou2024smolsearch, wang2025advancing} depend on constructing paired data using predefined similarity measures. Consequently, these methods either rely on costly annotations or are still tied to existing similarity tools.

Given these constraints, a compelling alternative is to leverage pretrained molecular representations that have already learned broad chemical knowledge from massive, unlabeled databases. By utilizing these foundational models, we can bypass the need for task-specific labels and instead extract general-purpose embeddings that capture the intrinsic structural features of the chemical space. While this concept has demonstrated good utility through finetuning for downstream tasks such as quantitative structure–activity relationship (QSAR) modeling \cite{Li2020,Ross2022,D5DD00348B,burns2026deeplearningfoundationmodels} and ADMET (Absorption, Distribution, Metabolism, Excretion, and Toxicity) prediction \cite{10.1093/nar/gkae298,doi:10.1021/acs.jcim.5c02106,xu2026smilesmambachemicalmambafoundation}, its application as a direct similarity function remains underexplored. Specifically, it lacks a comprehensive study of these embedding-derived similarities in guiding ligand-based virtual screening and generative molecular design. Moreover, it is critical to investigate whether these embedding-derived similarities can match or even exceed the utility of computationally intensive 3D alignment methods, offering a more scalable and efficient framework for navigating chemical space. To fill this gap, as shown in Figure \ref{fig:graphic_abstract}, this work investigates Pretrained Embedding Distance (PED) derived from two architecturally distinct molecular models: GeoDiff \cite{xu2022geodiff}, a diffusion-based model trained for molecular conformation generation, and MoLFormer \cite{Ross2022}, Transformer-based model trained on Simplified Molecular Input Line Entry System (SMILES) strings \cite{doi:10.1021/ci00057a005} using masked language modeling. Without any task-specific fine-tuning or explicit similarity supervision, we evaluate the utility of PED across three important tasks:
\begin{itemize}
    \item Correlation analysis between PED and traditional 3D alignment methods.
    \item Large-scale virtual screening on the challenging LIT-PCBA benchmark \cite{LIT-PCBA}.
    \item Reinforcement learning-based molecular generation including SMILES-based generation like REINVENT \cite{D3SC03781A} and synthesizable generation like SynFormer \cite{gao2025generative}.
\end{itemize}

Our results show that PED provides a competitive and flexible similarity signal across both virtual screening and molecular generation, suggesting that pretrained molecular embeddings can serve as a strong and general-purpose alternative to traditional similarity measures.

\section{Related Work}

\paragraph{Molecular Embedding-derived Similarity.}

Several studies use learned embedding space to design similarity or distance metric for molecular analog search and virtual screening. For example, DrugCLIP \cite{gao2023drugclip} reformulates virtual screening as a similarity matching problem between proteins and molecules, aligning protein pocket and molecule representations in a contrastive learning manner, which enables efficient large-scale screening without relying on explicit docking or affinity labels.  CHEESE (CHEmical Embeddings Search Engine) \cite{lvzivcavr2024cheese} accelerates virtual screening by embedding complex 3D and electrostatic similarities into a vector space, enabling billion-scale hit identification through rapid approximate nearest neighbor searches. S-MolSearch \cite{zhou2024smolsearch} is the first semi-supervised contrastive learning framework that leverages binding affinity signals for ligand-based virtual screening. It trains a 3D encoder on affinity-paired molecules, then uses it to generate similarity-based soft labels for unlabeled pairs, and train another encoder for large-scale retrieval. In addition, a prior pretrained molecular language model, MoLFormer \cite{Ross2022}, has shown that its embedding space is informative of chemical structural similarity. Specifically, based on the randomly sampled molecule pairs, their embedding similarities correlate with traditional fingerprint-based similarities, and the embedding Euclidean distances correlate with the atom numbers in maximum common subgraph. These studies suggest that embedding similarity can act as an operational measure for retrieval and structure-aware molecular comparison. However, most existing approaches rely on models that are explicitly trained for similarity learning. It remains unclear whether distances in embeddings from non-similarity-specific pretrained molecular models, can reliably capture chemically meaningful similarity, and whether they can further serve as a unified measure for drug discovery tasks.



\paragraph{RL-guided Molecular Generation.}

Reinforcement learning (RL) has been widely used for molecular generation, where the generator is iteratively updated to maximize the reward function, such as molecular properties \cite{Olivecrona2017,D3SC03781A}, binding affinity, and structure-aware similarity \cite{doi:10.1021/acs.jcim.0c01015} to target molecules. In this way, the model updates its policy to favor regions of chemical space that satisfy these complex requirements. For example, REINVENT \cite{D3SC03781A} is a SMILES-based RL framework that trains a generative agent from a pretrained prior and updates it through iterative sampling and scoring using an augmented likelihood that balances reward and prior constraints. LIMO \cite{eckmann2022limo} are graph-based generative models using RL to improve molecular properties in generation. MOFF \cite{wang2025improving} is a fragment-based reinforcement learning framework that uses docking score as rewards to generate both covalent and non-covalent inhibitors. In the context of ligand-based generative design, the selection of similarity score function directly decides the quality of generation. Fingerprint-based similarity may fail to capture higher-level structural or functional relationships, while docking-based and 3D similarity rewards are computationally expensive and difficult to scale. This challenge motivates customized reward design like embedding-based similarity as an alternative way for fast and approximated evaluation.


\section{Method}

\subsection{Pretrained Embedding Distance (PED)}

We define Pretrained Embedding Distance (PED) as a general similarity measure computed in the representation space of frozen pretrained molecular models. Given a pretrained molecular encoder $f(\cdot)$, a molecule represented by SMILES string $s$ is mapped to a vector embedding:
\begin{equation}
\mathbf{z} = f(s) \in \mathbb{R}^d
\end{equation}

Given two molecules $s_i$ and $s_j$ (e.g., a molecule and a reference ligand), their similarity is measured via a distance function $\mathrm{dist}$, such as cosine distance or Euclidean distance:
\begin{equation}
D(s_i, s_j) = \mathrm{dist}(f(s_i), f(s_j))
\end{equation}

We consider multiple representation modes: \textbf{2D}: embeddings derived from molecular graphs or SMILES. \textbf{3D}: embeddings derived from 3D conformations. \textbf{Concat}: concatenation after $\ell_2$ normalization. These variants allow us to analyze how different structural information contributes to embedding-space similarity.

\subsection{Instantiations with Pretrained Molecular Models}

We compute PED using two representative pretrained models: GeoDiff and MoLFormer, which provide 2D, 3D and sequence-based representations. The model settings are detailed in Appendix \ref{app:setting}.

\paragraph{GeoDiff.}

GeoDiff \cite{xu2022geodiff} is a diffusion-based model for molecular conformation generation. It models the distribution of molecular conformations by learning to reverse a noise perturbation process. Given a noisy conformation $\mathcal{C}_t$, GeoDiff predicts the noise term $\hat{\epsilon}_\theta$ using a dual-encoder architecture with two corresponding graph field networks (GFN). The dual encoder consists of: a 2D Graph Isomorphism Network (GIN) \cite{xu2019powerfulgraphneuralnetworks} encoding molecular topology, and a 3D equivariant SchNet \cite{NIPS2017_303ed4c6} encoding spatial structure. Overall, the model is trained with a denoising objective:
\begin{equation}
\mathcal{L}_{\text{GeoDiff}} = \sum_{t=1}^T \gamma_t \mathbb{E}_{\mathcal{C}_0, \epsilon} \|\epsilon - \hat{\epsilon}_\theta(\mathcal{C}_t, t)\|^2
\end{equation}

where $t$ is the index for diffusion steps, $\mathcal{C}_0$ is the ground truth conformation, $\gamma_t$ is a weight.

We use the pretrained model to extract embeddings from the dual-encoder components: (i) 2D embedding from the GIN encoder, (ii) 3D embedding from the SchNet encoder, and (iii) the concat embedding by concatenating both $\ell_2$ normalized representations. These embeddings are mean-pooled over atoms from the last layer to obtain fixed-length molecular representations.

\paragraph{MoLFormer.}

MoLFormer \cite{Ross2022} is a large-scale chemical language model pretrained on 1.1 billion SMILES sequences using a masked language modeling objective. For a given SMILES string, the model’s Transformer architecture generates a hidden representation for each individual token (atom or symbol). To derive a fixed-length molecular embedding from these token-level outputs, mean-pooling is applied across the last hidden states. This process aggregates the contextual information from the entire sequence into a global representation, capturing the latent chemical patterns and structural motifs learned from large-scale unlabeled data. The resulting high-dimensional vector serves as the basis for computing the corresponding PED.

\subsection{Applications of PED in Virtual Screening and Molecular Generation}

We evaluate PED on two representative drug discovery tasks: large-scale virtual screening and ligand-based generative molecular design.

\subsubsection{Virtual Screening}

Given a reference ligand and a library of candidate molecules, we compute embeddings for both and rank candidates based on PED in ascending order. Molecules with smaller distances are considered more similar to the reference and are prioritized. We evaluate PED on the LIT-PCBA benchmark, which consists of 15 targets, each with multiple reference ligands and a library of active and inactive candidate molecules. We follow the practice of \cite{LIT-PCBA,roshambo2} to use the enrichment factor at 1\% (EF1\%) as the metric, which measures the concentration of retrieved active compounds in the top-ranked subset:

\begin{equation}
\label{eq:ef}
\mathrm{EF}1\% = \frac{N_{\text{actives}}^{1\%}}{N^{1\%}} / \frac{N_{\text{actives}}}{N_{total}}
\end{equation}

To account for multiple reference ligands per target, we adopt a best-pooled evaluation strategy. For each candidate molecule $s_i$ and a set of reference ligands $\{s_r^{(j)}\}$, we compute the PED with respect to each reference and take the minimum distance:
\begin{equation}
D_{\text{pool}}(s_i) = \min_{j} D(s_i, s_r^{(j)})
\end{equation}

If the approach relies on similarity measures, a maximum similarity would be taken for best-pooling. The pooled distances are then used to rank all candidates, and EF1\% is computed based on the resulting ranking using Equation \ref{eq:ef}. This strategy allows each candidate molecule to give its best chance to match any of the known ligands for the target, better reflecting practical virtual screening settings where multiple reference ligands are available to search a chemical library \cite{roshambo2}.

\subsubsection{Molecular Generation}
Generative molecular design has emerged as a powerful paradigm for accelerating drug discovery, enabling \textit{in silico} exploration of vast chemical spaces to find molecules with specific structural and functional properties. In ligand-based design, this methodology is typically formulated as an iterative optimization loop centered on a reference template. With the goal of maximizing the desired properties, in each step, the base generative model samples a batch of candidate molecules, which are evaluated by a similarity or distance-based scoring function as reward signals. These rewards are then used to update the model, iteratively shifting the generative distribution toward the chemical space defined by the reference compound. In what follows, we introduce two representative generative models and evaluate how PED serves as an effective scoring function to guide the generation. 

\paragraph{REINVENT.}
REINVENT \cite{D3SC03781A} utilizes a SMILES-based chemical language model, pretrained on expansive molecular databases, as the foundation for generative design. In ligand-based molecular generation, it is as a reinforcement learning task where the pretrained model acts as a prior to ensure chemical validity. This agent model is finetuned to generate molecules toward the specific reference by optimizing an augmented objective, which increases the probability of high-scoring molecules while regularizing the agent toward the prior distribution. The agent network is then updated by minimizing the squared difference between its likelihood and this augmented likelihood.

We integrate PED into the REINVENT framework as a scoring component to guide molecular generation toward a reference ligand. A reverse sigmoid function is used to transform PED to a bounded reward in $[0, 1]$, where smaller distance yields higher reward (Equation \ref{eq:sigmoid}). Compared to cosine distance, raw Euclidean distance is unbounded and exhibits a larger dynamic range, providing a more informative reward signal. Therefore, we prioritize Euclidean PED in our generation experiments.

In practice, the final scoring function combines the PED-based score with additional constraint term that penalizes molecules containing undesirable substructures. Both components are weighted equally, ensuring that the model balances similarity optimization with chemical validity. Detailed experimental settings and hyperparameters are provided in Appendix \ref{app:setting}.

\paragraph{SynFormer.} Besides SMILES-based molecular generation, synthesizable molecular generation is another critical task in AI-aid drug discovery with real-world utility. It reformulates molecular design by first generating its sequential synthetic pathway and then assembling the final molecule. SynFormer \cite{gao2025generative} is a representative architecture, which utilizes a hybrid Transformer and diffusion framework and operates over a joint vocabulary of building block tokens and reaction tokens, generating a discrete sequence that encodes both the components and the assembly order of a synthesizable molecule. It is pretrained on a large collection of synthetic trajectories, learning a prior over chemically valid construction sequences before any task-specific optimization. During generation, SynFormer takes a reference SMILES as input and autoregressively decodes a building block-reaction sequence conditioned on the encoded input. It is further finetuned using a variant of the REINFORCE algorithm \cite{williams1992simple}, a type of RL approach, to guide its generation toward high-scoring molecules. 

\section{Experiments and Results}

\subsection{Evaluating PED for Virtual Screening}

\subsubsection{PED under Different Modes correlates with Corresponding 3D Similarities}

To validate whether pretrained embeddings capture meaningful chemical and spatial features, we first assess the correlation between PED and traditional 3D score that characterizes molecular similarity through a combination of shape (volumetric overlap) and color (pharmacophore matching).  As for the pretrained model used in PED, GeoDiff learns both topological information via a 2D GIN encoder and spatial structure via a 3D SchNet encoder, we hypothesize that its latent space implicitly aligns with these structural and chemical features. We choose AmpC dataset \cite{zhonglin} for this correction study, as pre-computed ROCS 3D scores for all molecules are public available. For this analysis, we sample 200k molecules from this dataset, uniformly across ROCS 3D combination similarity bins from 0 to 2 with an interval of 0.2. For each molecule, Color Tanimoto similarity, Shape Tanimoto similarity, and their balanced 50:50 combination score relative to a reference ligand are provided. We compute PED using MoLFormer and GeoDiff under different modes, and compare them with the corresponding ROCS similarities.

\begin{figure}[ht]
    \centering
    \includegraphics[width=1.\linewidth]{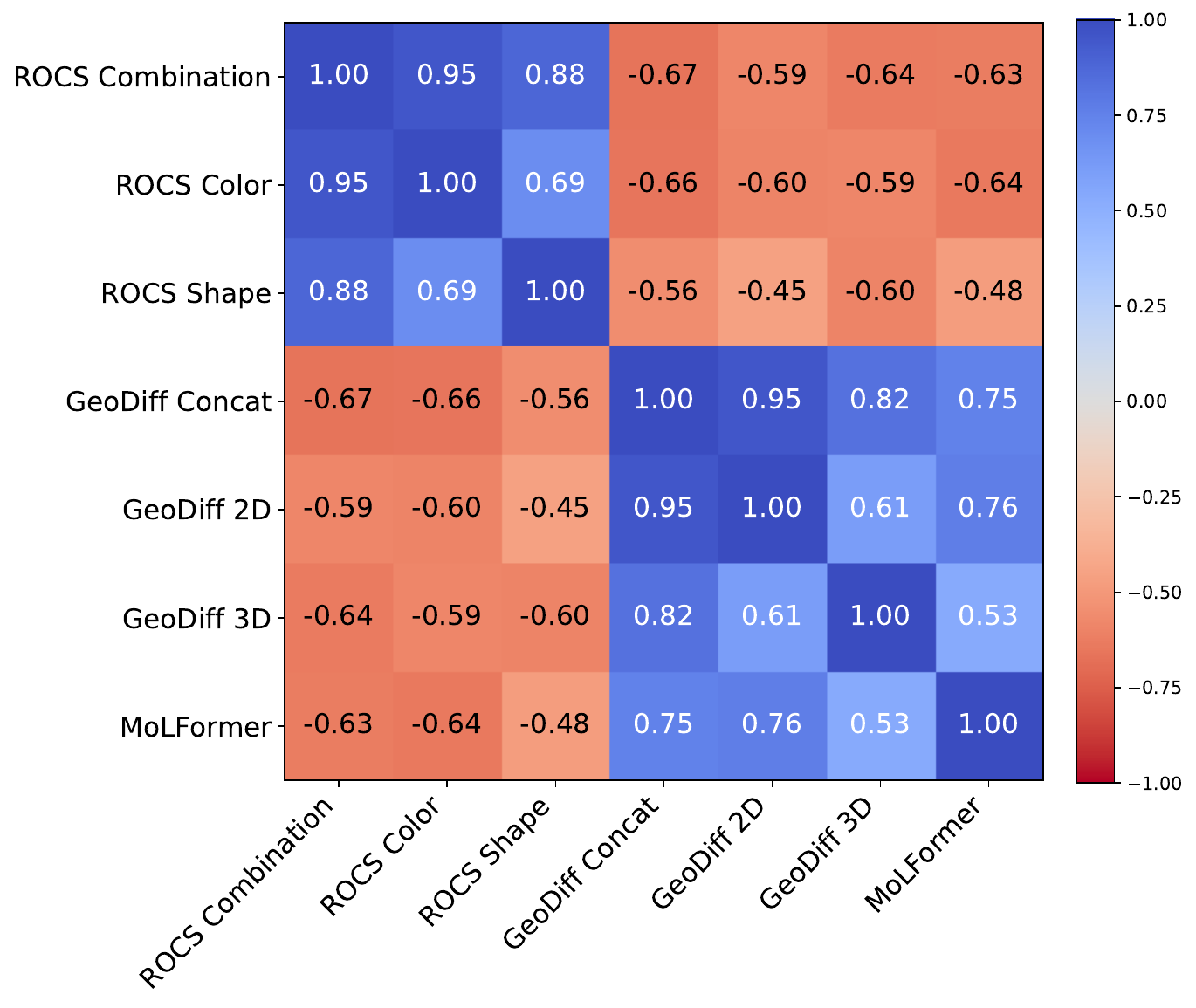}
    \caption{\textbf{Pearson correlation matrix across ROCS similarities, GeoDiff variant and MoLFormer cosine PEDs.} We focus on the top right corner to analyze the correlations between ROCS similarities and PEDs. As smaller distances correspond to higher similarity, the correlation sign is negative.}
    \label{fig:matrix_cosine}
\end{figure}

Figure \ref{fig:matrix_cosine} shows the Pearson correlations between each pair of all measures. The binned correlation plots for specific pairs can be found in Appendix \ref{app:correlation} (Figure \ref{fig:geodiff_2d_color} to \ref{fig:molformer_shape}). We observe that GeoDiff 2D PED highly correlates with ROCS color similarity ($r=-0.60$), and GeoDiff 3D PED shows comparable correlation with ROCS shape similarity ($r=-0.60$). It is important to note that these strong negative values indicate high alignment: as the traditional similarity score increases, the PED decreases. When considering mismatched pairs, correlations generally decrease, particularly for 2D PED (2D vs. shape: $r=-0.45$), whereas 3D PED remains relatively stable (3D vs. color: $r=-0.59$). Motivated by the fact that combined similarity often provides stronger signals, we concatenate normalized 2D and 3D embeddings and compute the joint PED. This leads to improved alignment with ROCS combination similarity ($r=-0.67$). Notably, GeoDiff 3D PED consistently demonstrates superior correlation across all ROCS metrics compared to its 2D counterpart. Besides, MoLFormer PED also exhibits relatively high alignment with ROCS combination similarity ($r=-0.63$), and exhibits higher correlation with color than shape ($r=-0.64$ vs.\ $-0.48$), which is consistent with its SMILES-based 2D characteristic.

\begin{figure*}[!t]
    \centering
    \includegraphics[width=1.\linewidth]{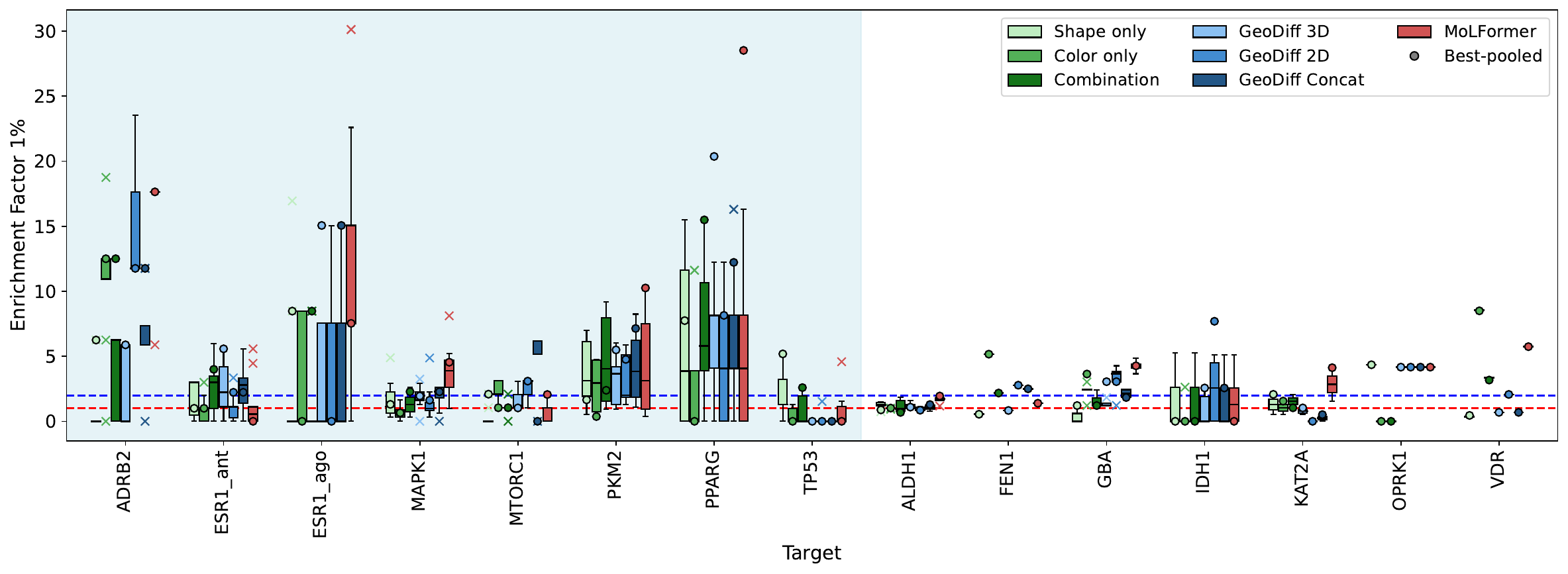}
    \caption{\textbf{LIT-PCBA virtual screening performance (EF1\%) obtained by ROSHAMBO2 similarities, GeoDiff and MoLFormer cosine PEDs.} Different modes of ROSHAMBO2 similarities are in green, while those for GeoDiff PEDs are in blue. Boxplots illustrate the distribution of EF1\% values across query ligands for each target. Circular markers indicate the best-pooled EF1\%, while cross markers denote the outliers of the boxplots. In case where there is only one reference ligand, the boxes would be a single line and the best-pooled circle overlaps with it. Targets highlighted in a blue background are those identified as favorable for 3D shape-based screening (EF1\% $>$ 2 with ROCS) in the original LIT-PCBA benchmark. Dashed horizontal red and blue lines correspond to EF1\% of 1 and 2, respectively.}
    \label{fig:lit_pcba_cosine}
\end{figure*}

\subsubsection{PED Outperforms Traditional Measures on Virtual Screening Benchmarks}

We benchmark PED against traditional 3D similarity measures on the LIT-PCBA virtual screening dataset. Figure \ref{fig:lit_pcba_cosine} presents the EF1\% distributions for three modes of 3D similarities computed by  ROSHAMBO2 \footnote{ROSHAMBO2 provides a robust, open-source alternative for 3D molecular similarity where commercial licensing of ROCS is restricted.} \cite{roshambo2}, three modes of GeoDiff cosine PEDs, and MoLFormer cosine PEDs. Details are reported in Tables \ref{tab:roshambo2_geodiff_complete_cosine} and \ref{tab:vs_comparison_mean_pooled_roshambo2_3D_cosine}.

For targets highlighted in blue (identified as favorable for 3D shape-based screening with EF1\% $>$ 2 using ROCS \cite{rocs,LIT-PCBA}), most PED variants achieve strong performance: 7 out of 8 targets obtain best-pooled EF1\% greater than 2. Notably, PED also performs well on less favorable targets, with 6 out of the remaining 7 targets exceeding the same threshold. Across all targets, MoLFormer cosine PED achieves the highest average mean and best-pooled EF1\% (4.53 $\pm$ 2.79 and 6.15), followed by 2D ECFP4 similarity (3.94 $\pm$ 2.43 and 4.83), as shown in Table \ref{tab:vs_comparison_mean_pooled_roshambo2_3D_cosine}.

Using Euclidean distance (Figure \ref{fig:lit_pcba_Euclidean}, Tables \ref{tab:roshambo2_geodiff_complete_Euclidean} and \ref{tab:vs_comparison_mean_pooled_roshambo2_3D_Euclidean}), the overall trends remain consistent. PED achieves best-pooled EF1\% greater than 2 on 7 out of 8 highlighted targets and 6 out of 7 remaining targets. MoLFormer Euclidean PED attains the highest average mean EF1\% (4.31 $\pm$ 3.11), while GeoDiff 3D Euclidean PED achieves the highest average best-pooled EF1\% (5.47), followed by MoLFormer (5.08).

Comparing different modes, ROSHAMBO2 achieves its highest average mean EF1\% under the color-only setting, while GeoDiff performs best under the 2D PED setting for both cosine and Euclidean distances. In contrast, for average best-pooled EF1\%, ROSHAMBO2 performs best under the combination mode, whereas GeoDiff achieves its highest values under the 3D PED setting.



\subsection{PED-Guided Molecular Generation}


To evaluate the impact of different embedding-based similarities in guiding molecular generation, we conduct a targeted case study using two generative frameworks, REINVENT and SynFormer, focused on the generation of novel analogs using a reference compound BMS-986195, a known Bruton's tyrosine kinase (BTK) inhibitor \cite{watterson2019discovery}. We compare three scoring regimes: ROSHAMBO2, which utilizes the experimental 3D coordinates from the reference’s crystallographic structure (PDB: 6O8I); transformed GeoDiff and MoLFormer Euclidean PEDs.

\paragraph{Training dynamics.}
Figure \ref{fig:reinvent_training} shows that REINVENT with all methods achieves stable convergence with high valid SMILES rates and exhibits distinct optimization behaviors. ROSHAMBO2 converges more slowly, whereas GeoDiff and MoLFormer-based PEDs lead to faster convergence of the reward. Meanwhile, SynFormer converges fast with a strong pretrained encoder for the reference compound (Figure \ref{fig:synformer_training}), so we only train it for fewer steps, resulting in a substantially smaller number of generated molecules compared to REINVENT (Table \ref{tab:generate_statistics}, Appendix \ref{app:statistics_mols}). Therefore, we focus on comparing different scoring methods within each framework. We also observe significant differences in computational efficiency. 3D-based scoring via ROSHAMBO2 exhibits the slowest convergence and highest computational overhead due to required conformer generation. In contrast, PED-based methods provide a substantial speed-up. According to Table~\ref{tab:generate_statistics}, in the REINVENT experiments, GeoDiff and MoLFormer-based PEDs achieve a 1.5$\times$ and 3.3$\times$ speed-up, respectively, compared to the ROSHAMBO2 baseline. These trends are consistently observed in the SynFormer experiments, where both GeoDiff and MoLFormer approximately double the sampling speed ($\sim$2$\times$ faster) relative to ROSHAMBO2. These results underscore the computational advantage of leveraging learned representations over explicit spatial optimization for speeding up molecular generation.

\begin{figure}[ht]
    \centering
    \includegraphics[width=1.\linewidth]{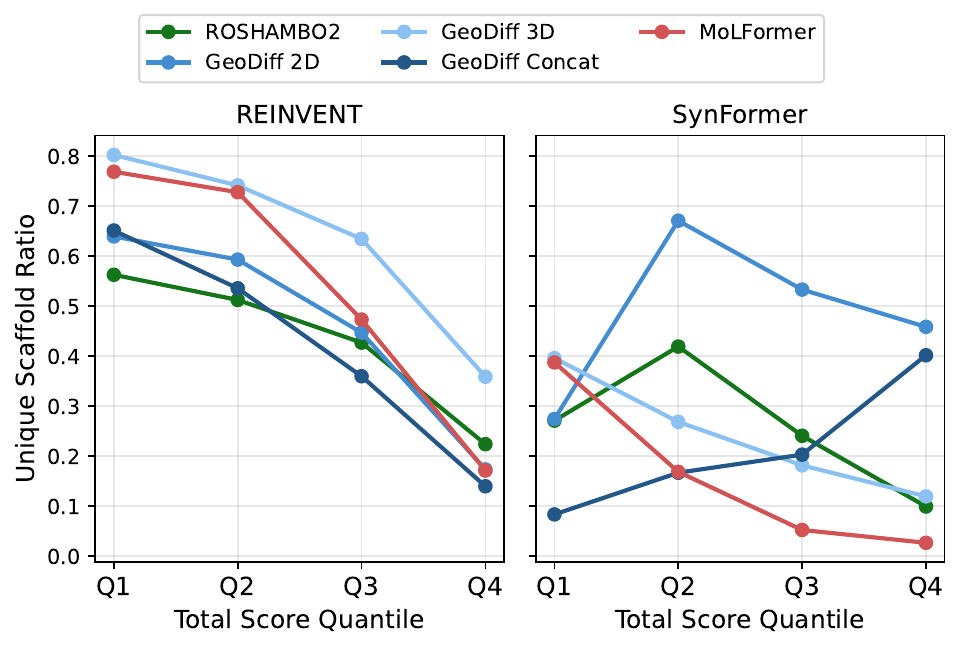}
    \caption{\textbf{Scaffold diversity across total score quantiles.} For each method, molecules are grouped into four equal-frequency bins based on total score (Q1–Q4, from low to high). The y-axis shows the ratio of unique ring-linker scaffold to the total number of molecules within each bin.}
    \label{fig:scaffold_quantile}
\end{figure}

\paragraph{Scaffold diversity of generated molecules.}
Figure \ref{fig:scaffold_quantile} shows that scaffold diversity of REINVENT-generated molecules consistently decreases as total score increases across all methods, indicating that higher-scoring molecules are more structurally concentrated. This trend is most pronounced for MoLFormer and GeoDiff Concat, suggesting stronger convergence. MoLFormer and GeoDiff 3D maintain comparatively higher diversity in the early stages, exhibiting strong exploration in low-score regions, while MoLFormer shifts toward more focused exploitation in high-score regions. In contrast, SynFormer exhibits a different pattern. GeoDiff 2D achieves the highest scaffold diversity across most quantiles, particularly in Q2, indicating strong exploration behavior. The scaffold diversity of GeoDiff Concat increases toward higher-score regions. However, other methods show a decrease trend of scaffold diversity along with the total score, which are similar to REINVENT.

Table \ref{tab:top_5000_scaffold} further summarizes the unique scaffold count and ratio within the top-5,000 molecules. This corresponds to the highest score quantile (Q4) of REINVENT, while it covers most of the SynFormer-generated unique molecules due to the smaller sample size. Within REINVENT, GeoDiff 3D achieves the highest scaffold diversity (35.16\%), followed by MoLFormer (12.84\%), while other methods remain lower. In SynFormer, GeoDiff 2D shows the highest scaffold ratio (46.36\%), substantially exceeding other methods, but MoLFormer produces the lowest diversity (5.04\%). Despite differences in scale, both frameworks indicate that embedding-based scoring can significantly affect the diversity of high-scoring molecules.

\begin{table}[ht]
    \centering
    \caption{Unique ring-linker scaffold and its ratio over top-5,000 high-score generated molecules.}
    \resizebox{0.5\textwidth}{!}{
    \begin{tabular}{llcc}
    \toprule
         Model & Method & Unique scaffold & Unique scaffold ratio ($\uparrow$) \\
    \midrule
    \multirow{5}{*}{REINVENT}
         & ROSHAMBO2      & 397   & 7.94\%  \\
         & GeoDiff 2D     & 379   & 7.58\%  \\
         & GeoDiff 3D     & 1,758 & \textbf{35.16\%} \\
         & GeoDiff Concat & 301   & 6.02\%  \\
         & MoLFormer      & 642   & 12.84\% \\
    \midrule
    \multirow{5}{*}{SynFormer}
         & ROSHAMBO2      & 1,041   & 20.82\%  \\
         & GeoDiff 2D     & 2,318   & \textbf{46.36\%}  \\
         & GeoDiff 3D     & 624     & 12.48\% \\
         & GeoDiff Concat & 1,046   & 20.92\%  \\
         & MoLFormer      & 252     & 5.04\% \\
    \bottomrule
    \end{tabular}
    }
    \label{tab:top_5000_scaffold}
\end{table}


\paragraph{Drug-likeness of top-5,000 generated molecules.}
Table \ref{tab:physchem_top5000} and \ref{tab:druglikeness_percent} in Appendix \ref{app:drug_likeness} show drug-likeness statistics of top-5,000 high-score molecules. REINVENT consistently achieves strong compliance with good drug-like property ranges across all methods, which can be attributed to the unfavorable substructure filters in the reward function. In contrast, SynFormer exhibits much larger variation in physicochemical properties due to its building-block-based generation process, which allows more substantial structural modifications. This effect is particularly evident for methods with higher scaffold diversity, such as GeoDiff 2D and GeoDiff Concat, where molecules frequently fall outside desirable ranges in molecular weight, TPSA, LogP, and QED. These results highlight a trade-off between structural diversity and drug-likeness across different generative frameworks.



\paragraph{Binding affinity prediction of top-score generated molecules by Boltz-2.}
To validate the biological relevance of our generative results, we select the top-1,000 molecules ranked by the total score for each method to assess their predicted potency against the BTK target using Boltz-2 \cite{Passaro2025.06.14.659707}. Boltz-2 is a particularly reliable evaluator for this task as it has demonstrated high fidelity on the kinase family \cite{Passaro2025.06.14.659707, cecchini2026rise}. We focus on the pIC50 metric, where a higher value indicates lower IC50 concentration and thus higher potency, to assess whether molecules favored by different reward signals are also predicted to exhibit strong predicted inhibition. To reduce bias from over-represented scaffolds, we further construct a scaffold-balanced subset by grouping the top-1,000 molecules by ring-linker scaffold and uniformly sampling up to 500 molecules per method.

\begin{figure}[t]
    \centering
    \includegraphics[width=1.\linewidth]{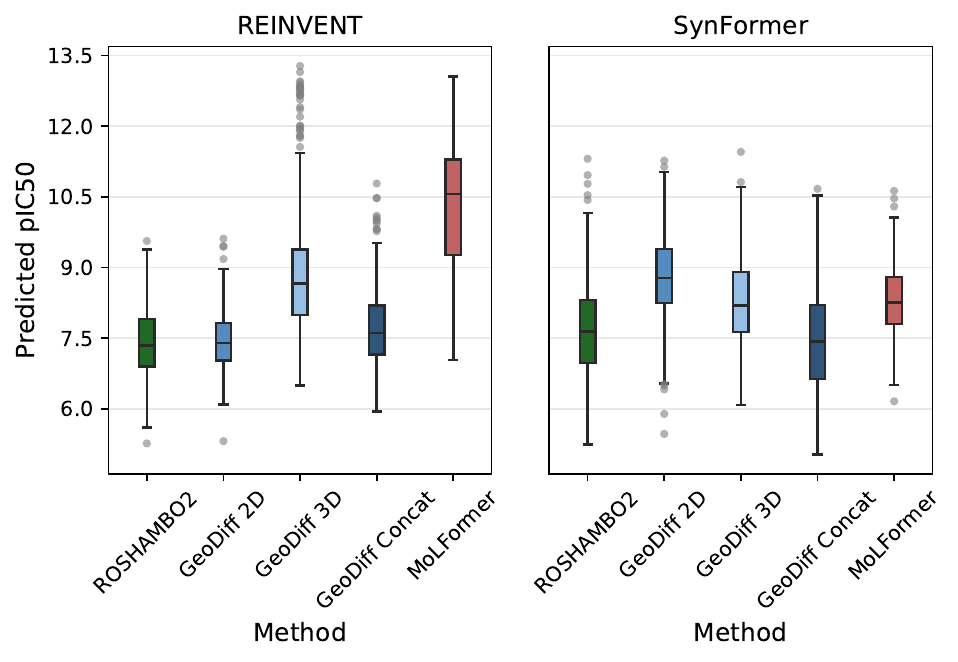}
    \caption{\textbf{Predicted pIC50 distribution of scaffold-balanced top-ranked generated molecules.} For each method, the top-1,000 molecules are first grouped by scaffold. We then uniformly sample 500 molecules across scaffolds for each method.}
    \label{fig:pic50_boltz_500}
\end{figure}

As shown in Figure~\ref{fig:pic50_boltz_500} and Table~\ref{tab:pic50_500_stats}, the ranking of reward signals differs across the two generative frameworks. For REINVENT, MoLFormer yields the highest predicted pIC50 distribution ($10.27 \pm 1.34$, effect size $\Delta=0.92$), followed by GeoDiff 3D ($8.83 \pm 1.29$, $\Delta=0.71$). GeoDiff Concat shows only moderate improvement ($7.70 \pm 0.79$, $\Delta=0.20$), while ROSHAMBO2 and GeoDiff 2D remain lower and close to each other ($7.40 \pm 0.69$ vs. $7.44 \pm 0.60$, $\Delta=0.04$). For SynFormer, the overall differences are less significant, but embedding-based rewards still outperform ROSHAMBO2 in most cases. GeoDiff 2D achieves the highest predicted pIC50 ($8.81 \pm 0.87$, $\Delta=0.61$), followed by MoLFormer ($8.31 \pm 0.74$, $\Delta=0.39$) and GeoDiff 3D ($8.28 \pm 0.90$, $\Delta=0.34$), whereas GeoDiff Concat performs slightly worse than ROSHAMBO2.

\begin{table}[ht]
\centering
\caption{\textbf{Predicted pIC50 statistics of scaffold-balanced top-500 molecules for REINVENT and SynFormer.} Cliff’s delta is reported as an effect size, measuring how often the predicted pIC50 values from a PED-based method exceed those from ROSHAMBO2.}
\resizebox{0.5\textwidth}{!}{
\begin{tabular}{llccc}
\toprule
Model & Method & Predicted pIC50 ($\uparrow$) & Effect size \\
\midrule
\multirow{5}{*}{REINVENT} & ROSHAMBO2 & 7.40 $\pm$ 0.69 &  — \\
& GeoDiff 2D & 7.44 $\pm$ 0.60 & 0.04 \\
& GeoDiff 3D & 8.83 $\pm$ 1.29 & 0.71 \\
& GeoDiff Concat & 7.70 $\pm$ 0.79 & 0.20 \\
& MoLFormer & \textbf{10.27 $\pm$ 1.34} & \textbf{0.92} \\
\midrule
\multirow{5}{*}{SynFormer} & ROSHAMBO2 & 7.69 $\pm$ 1.04  & — \\
& GeoDiff 2D & \textbf{8.81 $\pm$ 0.87} & \textbf{0.61} \\
& GeoDiff 3D & 8.28 $\pm$ 0.90 & 0.34 \\
& GeoDiff Concat & 7.48 $\pm$ 1.06 & $-$0.11 \\
& MoLFormer & 8.31 $\pm$ 0.74 & 0.39 \\
\bottomrule
\end{tabular}
}
\label{tab:pic50_500_stats}
\end{table}

Overall, these results suggest that molecules generated with the guidance of PED-based reward signals are more likely to bind the target protein pocket, although the strongest PED mode varies with the generative framework. Specifically, REINVENT benefits most from MoLFormer and GeoDiff 3D, whereas SynFormer performs well with GeoDiff 2D and MoLFormer. The top-5 generated molecules ranked by total score with predicted pIC50 are shown in Appendix Figure~\ref{fig:top5_reinvent} and \ref{fig:top5_synformer}.

\section{Conclusion}

This study establishes a connection between similarity-based virtual screening and molecular generation via pretrained embedding distance (PED). Our findings show that PEDs derived from models like GeoDiff and MoLFormer exhibit strong correlations with industry-standard 3D similarity metrics, suggesting that pretrained molecular encoders implicitly capture rich spatial and pharmacophoric information without requiring explicit 3D alignment. This signal performs effectively across both virtual screening and reinforcement learning-based generation, providing a reliable ranking function for the challenging LIT-PCBA benchmark and a continuous reward signal for frameworks like REINVENT and SynFormer. Most notably, PED offers a substantial leap in computational efficiency, achieving up to a 3.3× speed-up in generative sampling by bypassing the need for expensive conformer generation and iterative spatial optimization.

\section*{Impact Statement}
This paper demonstrates that pretrained molecular embeddings can be repurposed as efficient similarity measurement to significantly accelerate early-stage drug discovery.  While this work focuses on a BTK inhibitor case study in molecular generation task and evaluates the generation compound using Boltz-2, the biological efficacy and safety of the generated compounds are still not wet-lab validated. Future work involving experimental synthesis and assays is required to confirm the therapeutic potential of the identified leads.

\bibliography{example_paper}
\bibliographystyle{icml2026}

\newpage
\appendix
\onecolumn

\renewcommand{\thefigure}{S\arabic{figure}}
\renewcommand{\thetable}{S\arabic{table}}
\renewcommand{\theequation}{S\arabic{equation}}
\setcounter{figure}{0}
\setcounter{table}{0}
\setcounter{equation}{0}




\section{Experiment Setting and Hyperparameters}
\label{app:setting}

\paragraph{Pretrained Molecular Models.}
We used GeoDiff \cite{xu2022geodiff} and MoLFormer \cite{Ross2022} models as two representative pretrained molecular models to compute PED. GeoDiff was pretrained on the GEOM-Drugs \cite{Axelrod2022} dataset, and MoLFormer was the XL version (\url{https://huggingface.co/ibm-research/MoLFormer-XL-both-10pct}) pretrained on 1.1B molecules from ZINC and PubChem (10\% of both datasets). Both 2D GIN and 3D SchNet encoders of GeoDiff generate 128-length embeddings, the size of the concatenated embedding from the dual encoder is 256. The size of MoLFormer's embedding is 768.

\paragraph{SMILES-based Generative Model.}
We used REINVENT 3.2 (\url{https://github.com/MolecularAI/Reinvent}) with a pretrained prior and agent initialized from the same checkpoint, and optimized the agent for 10,000 RL steps with batch size 128 and learning rate $1\times10^{-4}$. Each run generates 1,280,000 molecules (including duplicates). The scoring function was a equally weighted sum of two components—custom substructure alerts and sigmoid-transformed PED. An inception mechanism (memory size 100, sample size 10) was used to replay high-scoring molecules. The reward scaling parameter was set to 128, and a margin threshold of 50 was used during training.

\paragraph{Synthesizable Generative Model.}
For SynFormer, we used the default model architecture and pretrained weights provided by the authors' GitHub repository (\url{https://github.com/wenhao-gao/synformer}). This method operated over a synthesizable chemical space defined by 115 reaction templates and 223,244 commercially available building blocks sourced from the Enamine US in stock \cite{enamine}, covering an estimated space of $>10^{60}$ molecules. For RL finetuning, we also followed the original paper with minimal modification. Specifically, we retained the frozen prior model and employed a replay buffer that sampled 128 saved molecules at each step using a weighting mechanism over accumulated scores to stabilize the gradient updates. The augmented likelihood scaling factor was set to $\sigma = 500$ and the Adam optimizer was used with a learning rate of $1 \times 10^{-4}$, both matching the original paper's settings. We used a batch size of 128 and the maximal training step of 100, resulting in a total of 12,800 generated molecules per run (including duplicates).

\paragraph{Reverse Sigmoid Function.} Given a generated molecule $s_i$ and a reference molecule $s_r$, the distance between them in embedding space is transformed into a reward via a reverse sigmoid function:

\begin{equation}
\label{eq:sigmoid}
R(s_i) = \frac{1}{1 + 10^{\,10k \cdot \frac{D(s_i, s_r) - m}{h - l}}}
\end{equation}

where $k$ adjusts the sharpness of the reward, $m = \frac{h + l}{2}$ is the midpoint, and $l$ and $h$ denote the lower and upper bounds of the distance range, respectively. Specifically, Table \ref{tab:ped_hyperparams} displays the hyperparameters of PED sigmoid transformation in different distance modes, which are applied to the above two generative models as a scoring function component.

\begin{table}[ht]
\centering
\caption{Hyperparameters for the sigmoid transformation used to convert different Euclidean PEDs into reward signals.}
\label{tab:ped_hyperparams}
\begin{tabular}{lccc}
\toprule
Distance mode & $l$ (low) & $h$ (high) & $k$ \\
\midrule
GeoDiff 2D      & 0.1 & 1.2 & 0.25 \\
GeoDiff 3D      & 1   & 10  & 0.25 \\
GeoDiff Concat  & 0.1 & 0.58  & 0.25 \\
MoLFormer       & 5   & 17  & 0.25 \\
\bottomrule
\end{tabular}
\end{table}

Experiments were conducted in an HPC server by using an NVIDIA A100 GPU with 80GB of memory. The datasets we used are public and available in \url{https://drugdesign.unistra.fr/LIT-PCBA/} and \url{https://zenodo.org/records/10277786}. The code repository is available in GitHub: \url{https://github.com/molecularinformatics/PED}.

\section{Correlation Results of AmpC-ROCS Similarities and Embedding Distance}
\label{app:correlation}

\begin{figure}[ht]
    \centering
    \includegraphics[width=0.9\linewidth]{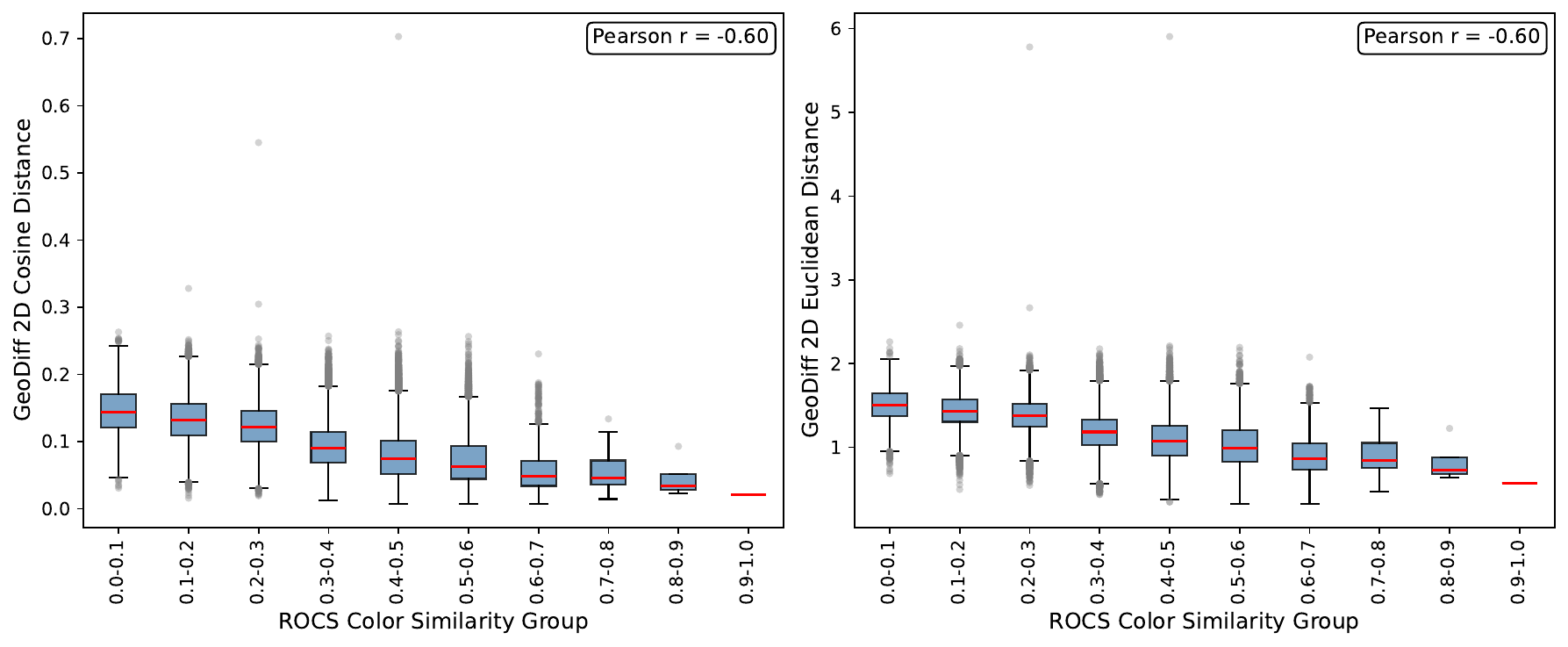}
    \caption{GeoDiff 2D cosine and Euclidean PED vs. ROCS ColorTanimoto.}
    \label{fig:geodiff_2d_color}
\end{figure}

\begin{figure}[ht]
    \centering
    \includegraphics[width=0.9\linewidth]{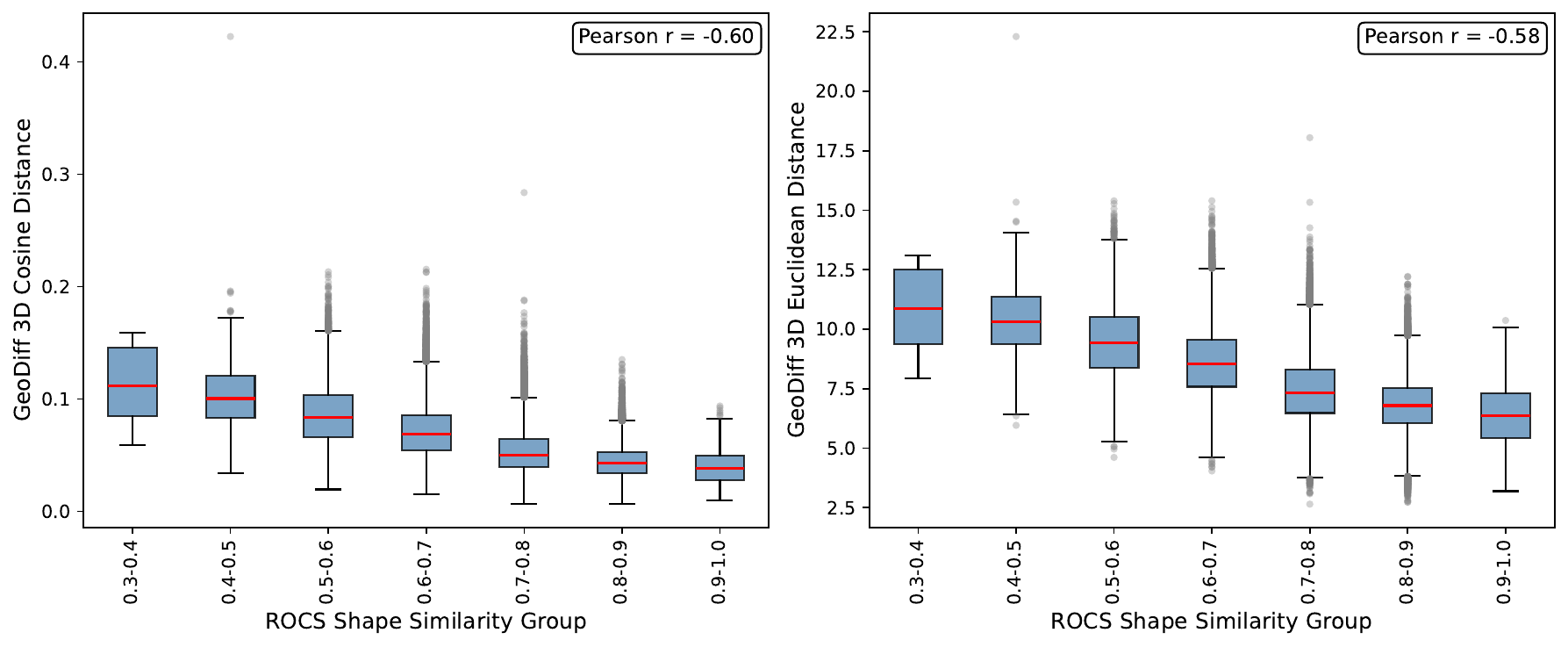}
    \caption{GeoDiff 3D cosine and Euclidean PED vs. ROCS ShapeTanimoto.}
    \label{fig:geodiff_3d_shape}
\end{figure}

\begin{figure}[ht]
    \centering
    \includegraphics[width=0.9\linewidth]{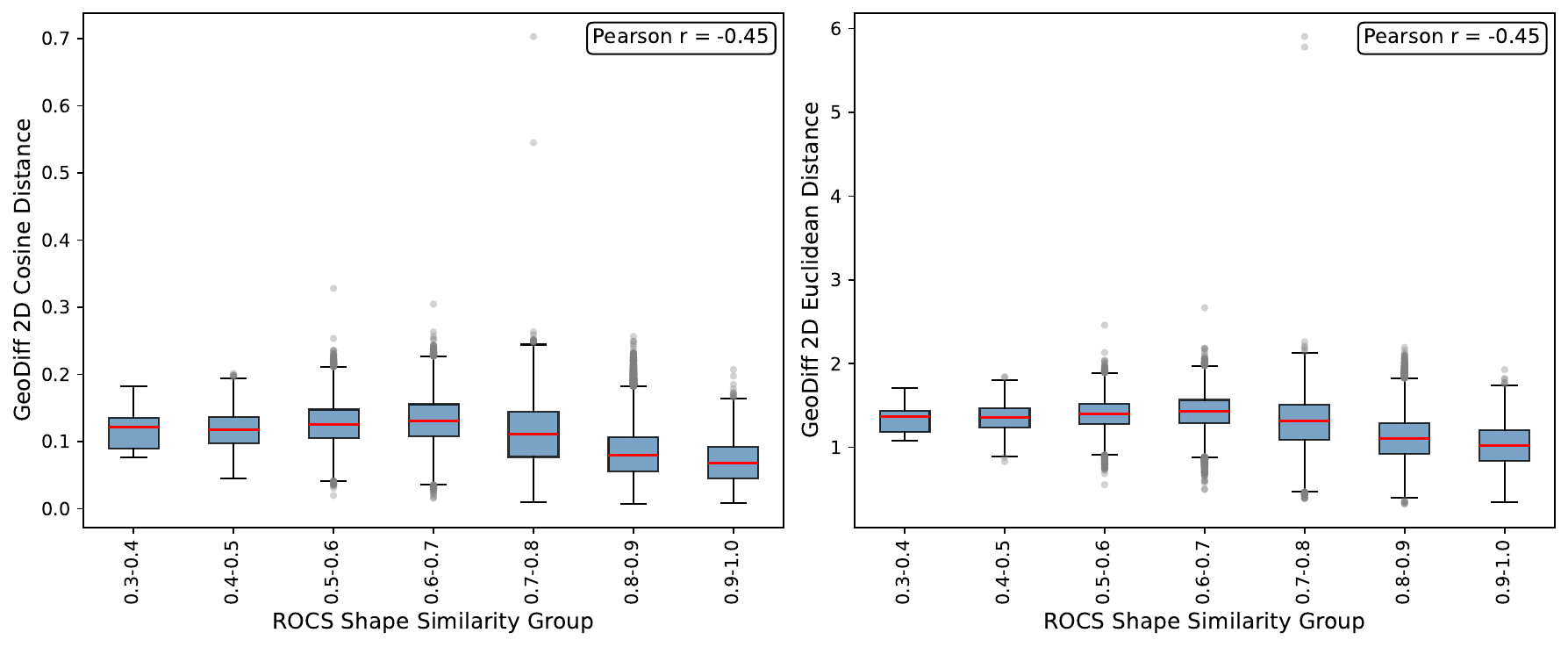}
    \caption{GeoDiff 2D cosine and Euclidean PED vs. ROCS ShapeTanimoto.}
    \label{fig:geodiff_2d_shape}
\end{figure}

\begin{figure}[ht]
    \centering
    \includegraphics[width=0.9\linewidth]{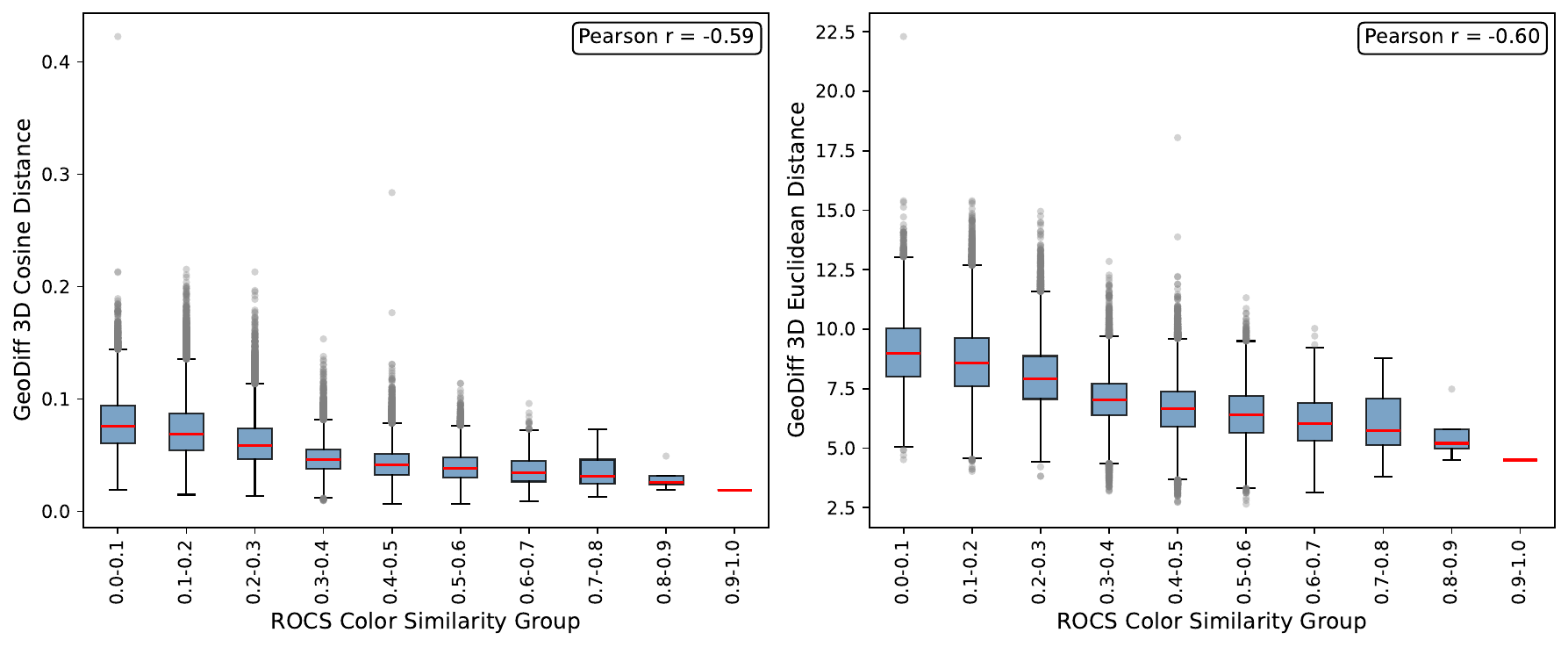}
    \caption{GeoDiff 3D cosine and Euclidean PED vs. ROCS ColorTanimoto.}
    \label{fig:geodiff_3d_color}
\end{figure}

\begin{figure}[ht]
    \centering
    \includegraphics[width=0.9\linewidth]{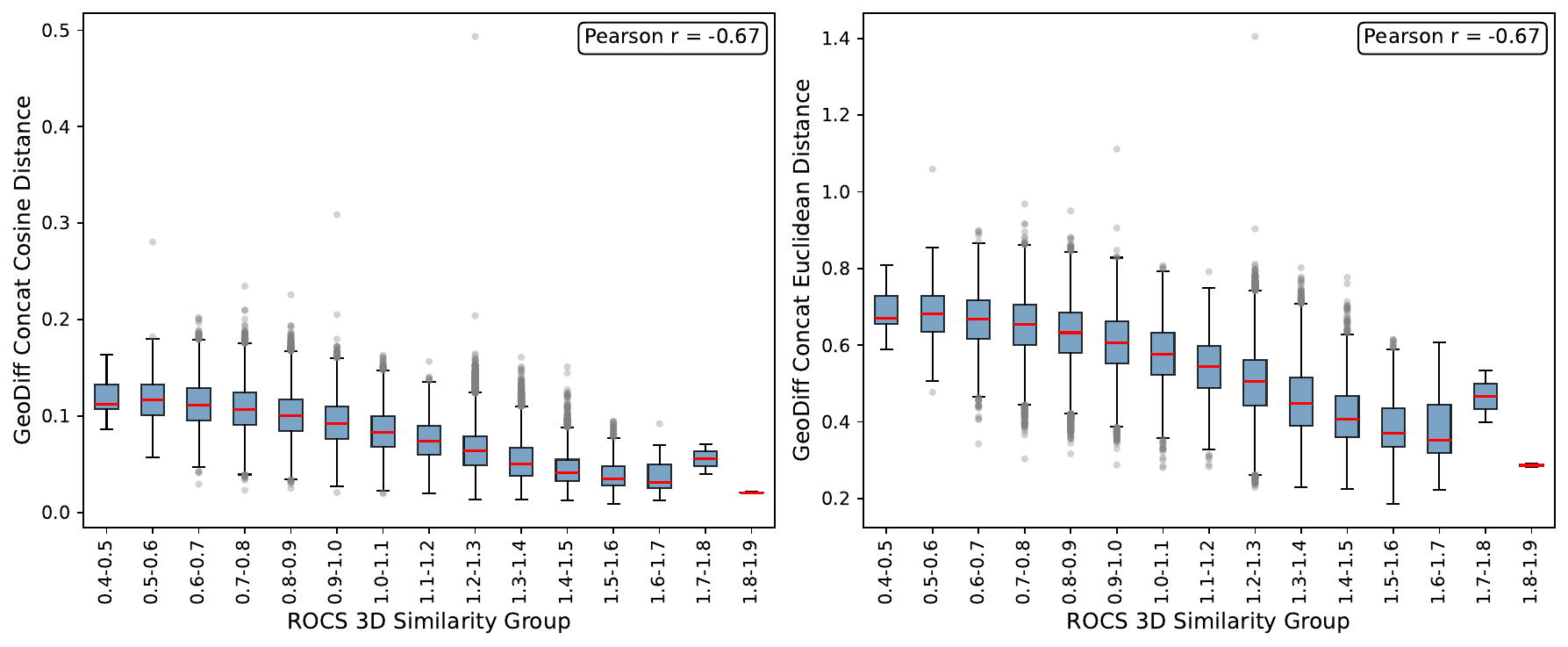}
    \caption{GeoDiff concatenated cosine and Euclidean PED vs. ROCS combination.}
    \label{fig:geodiff_concat_comb}
\end{figure}

\begin{figure}[ht]
    \centering
    \includegraphics[width=0.9\linewidth]{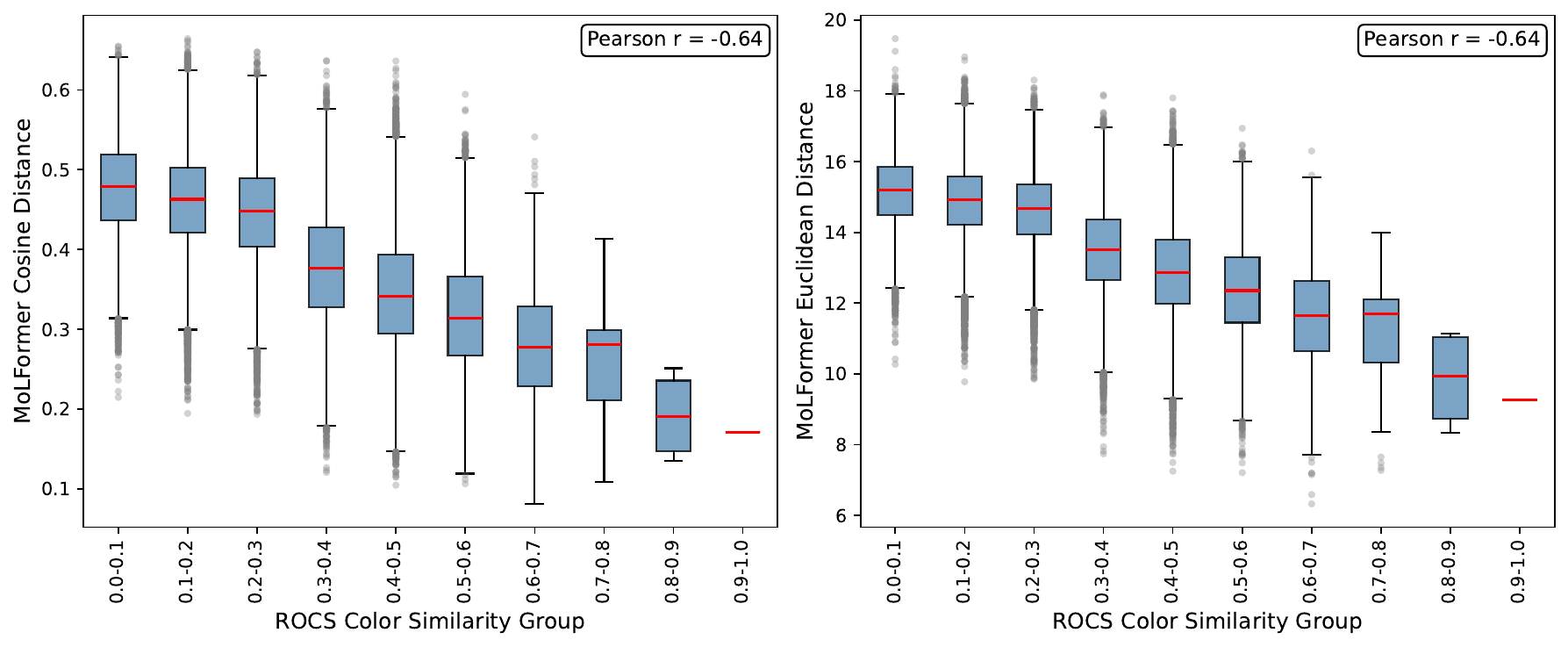}
    \caption{MoLFormer cosine and Euclidean PED vs. ROCS ColorTanimoto.}
    \label{fig:molformer_color}
\end{figure}

\begin{figure}[ht]
    \centering
    \includegraphics[width=0.9\linewidth]{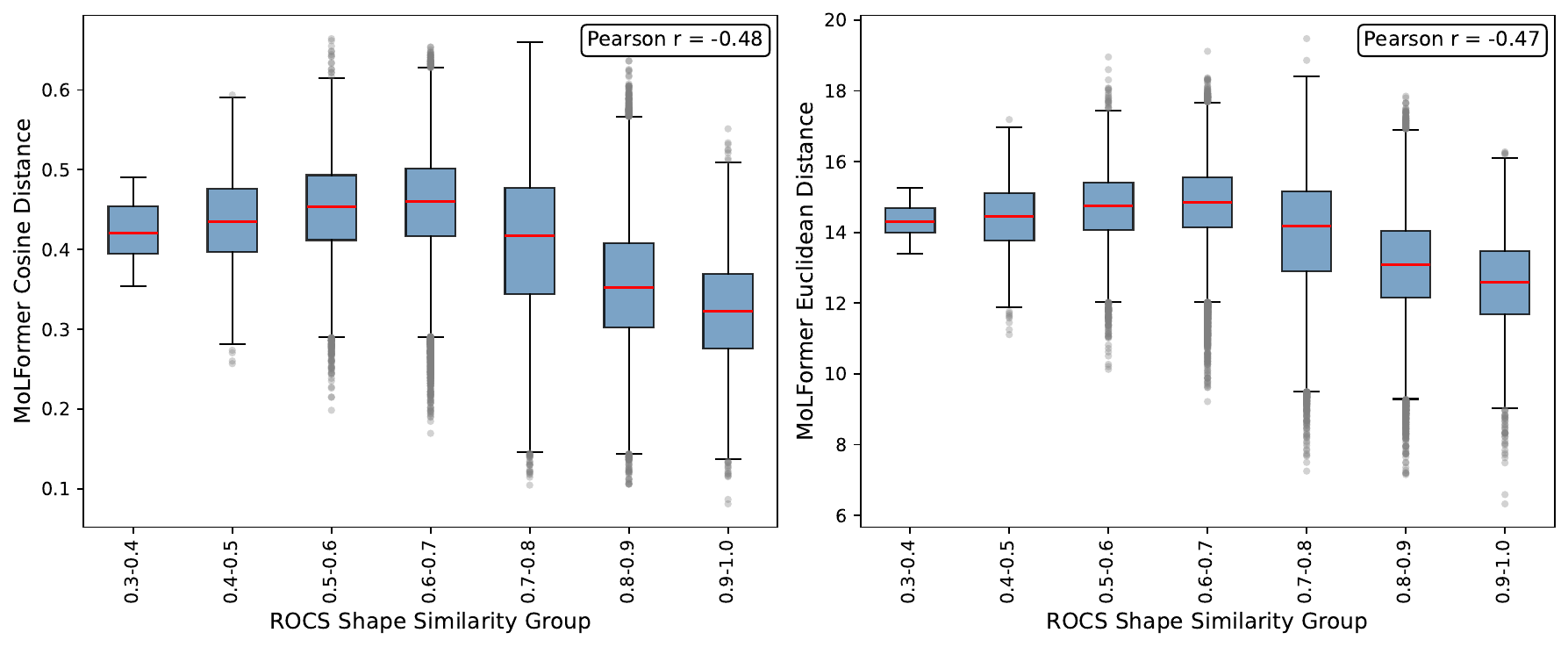}
    \caption{MoLFormer cosine and Euclidean PED vs. ROCS ShapeTanimoto.}
    \label{fig:molformer_shape}
\end{figure}

\clearpage

\section{LIT-PCBA Benchmark and Virtual Screening Results on EF1\%}



We used a challenging virtual screening dataset, LIT-PCBA \cite{LIT-PCBA}, to benchmark PED. However, Huang et al. \yrcite{huang2025dataleakageredundancylitpcba} found that the provided original reference ligands in the .mol2 files often lack bond orders, impeding accurate structure reconstruction and yielding incorrect SMILES. We then used their corrected canonical SMILES for all the reference ligands.

The EF1\% results of ROSHAMBO2 are provided by its original paper \cite{roshambo2}. 2D ECFP4 and ROCS 3D similarities and molecular docking results are from the original LIT-PCBA benchmark \cite{LIT-PCBA}.

\begin{figure}[ht]
    \centering
    \includegraphics[width=1.\linewidth]{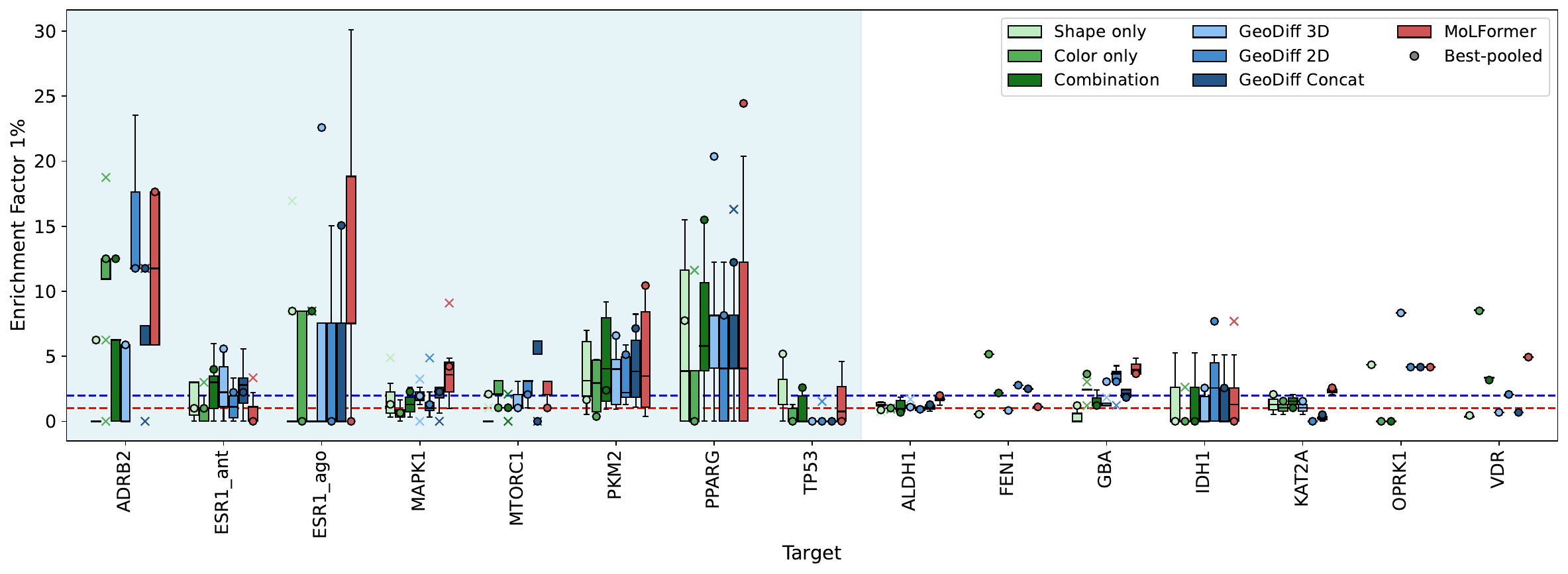}
    \caption{\textbf{LIT-PCBA virtual screening performance (EF1\%) obtained by ROSHAMBO2 similarities, GeoDiff and MoLFormer Euclidean PEDs.} Different modes of ROSHAMBO2 similarities are in green, while those for GeoDiff PEDs are in blue. Boxplots illustrate the distribution of EF1\% values across query ligands for each target. Circular markers indicate the best-pooled EF1\%, while cross markers denote the outliers of the boxplots. In case where there is only one reference ligand, the boxes would be a single line and the best-pooled circle overlaps with it. Targets highlighted in a blue background are those identified as favorable for 3D shape-based screening (EF1\% $>$ 2 with ROCS) in the original LIT-PCBA benchmark. Dashed horizontal red and blue lines correspond to EF1\% of 1 and 2, respectively.}
    \label{fig:lit_pcba_Euclidean}
\end{figure}

\begin{table}[ht]
\caption{\textbf{Complete virtual screening performance (EF1\%) obtained by different modes of ROSHAMBO2 similarity and GeoDiff cosine PED.} The best average mean and best-pooled EF1\% values for each approach are shown in \textbf{bold}.}
\centering
\resizebox{\textwidth}{!}{
\begin{tabular}{l cc | cc | cc | cc | cc | cc }
\toprule
& \multicolumn{2}{c}{ROSHAMBO2 color} & \multicolumn{2}{c}{ROSHAMBO2 shape} & \multicolumn{2}{c}{ROSHAMBO2 combination} & \multicolumn{2}{c}{GeoDiff 2D} & \multicolumn{2}{c}{GeoDiff 3D} & \multicolumn{2}{c}{GeoDiff concat} \\
\cmidrule(lr){2-3} \cmidrule(lr){4-5} \cmidrule(lr){6-7} \cmidrule(lr){8-9} \cmidrule(lr){10-11} \cmidrule(lr){12-13}
Target & Mean $\pm$ SD & Best-pooled & Mean $\pm$ SD & Best-pooled & Mean $\pm$ SD & Best-pooled & Mean $\pm$ SD & Best-pooled & Mean $\pm$ SD & Best-pooled & Mean $\pm$ SD & Best-pooled \\
\midrule
ADRB2 & 10.94 $\pm$ 5.54 & 12.50 & 0.78 $\pm$ 2.21 & 6.25 & 3.91 $\pm$ 3.24 & 12.50 & 14.70 $\pm$ 4.45 & 11.76 & 2.21 $\pm$ 3.04 & 5.88 & 6.62 $\pm$ 3.77 & 11.76 \\
ALDH1 & 0.98 $\pm$ 0.04 & 1.02 & 1.23 $\pm$ 0.30 & 0.88 & 1.18 $\pm$ 0.52 & 0.68 & 0.93 $\pm$ 0.18 & 0.85 & 1.20 $\pm$ 0.27 & 1.09 & 1.08 $\pm$ 0.25 & 1.29 \\
ESR1\_ago & 3.03 $\pm$ 4.21 & 0.00 & 1.82 $\pm$ 4.91 & 8.47 & 1.82 $\pm$ 3.61 & 8.47 & 4.02 $\pm$ 6.28 & 0.00 & 3.01 $\pm$ 3.82 & 15.06 & 4.02 $\pm$ 4.82 & 15.06 \\
ESR1\_ant & 0.73 $\pm$ 0.88 & 1.00 & 1.67 $\pm$ 1.34 & 1.00 & 2.53 $\pm$ 2.06 & 4.00 & 1.11 $\pm$ 0.98 & 2.23 & 2.39 $\pm$ 1.85 & 5.57 & 2.71 $\pm$ 1.56 & 2.23 \\
FEN1 & 5.16 & 5.16 & 0.54 & 0.54 & 2.17 & 2.17 & 2.78 & 2.78 & 0.83 & 0.83 & 2.50 & 2.50 \\
GBA & 2.30 $\pm$ 0.66 & 3.64 & 0.36 $\pm$ 0.54 & 1.21 & 1.58 $\pm$ 0.54 & 1.21 & 3.20 $\pm$ 1.35 & 3.05 & 1.37 $\pm$ 0.30 & 3.05 & 2.13 $\pm$ 0.35 & 1.83 \\
IDH1 & 0.20 $\pm$ 0.73 & 0.00 & 1.01 $\pm$ 1.71 & 0.00 & 1.01 $\pm$ 1.71 & 0.00 & 2.20 $\pm$ 2.22 & 7.69 & 0.73 $\pm$ 1.20 & 2.56 & 1.28 $\pm$ 1.67 & 2.56 \\
KAT2A & 1.04 $\pm$ 0.73 & 1.55 & 1.30 $\pm$ 1.10 & 2.07 & 1.55 $\pm$ 0.73 & 1.04 & 0.00 $\pm$ 0.00 & 0.00 & 0.77 $\pm$ 0.36 & 1.03 & 0.26 $\pm$ 0.36 & 0.52 \\
MAPK1 & 0.60 $\pm$ 0.48 & 0.65 & 1.56 $\pm$ 1.31 & 1.30 & 1.35 $\pm$ 0.67 & 2.28 & 1.42 $\pm$ 1.28 & 1.62 & 1.95 $\pm$ 0.86 & 1.95 & 1.95 $\pm$ 0.89 & 2.27 \\
MTORC1 & 2.37 $\pm$ 0.82 & 1.04 & 0.28 $\pm$ 0.67 & 2.09 & 1.04 $\pm$ 0.47 & 1.04 & 2.47 $\pm$ 0.92 & 3.09 & 1.65 $\pm$ 0.92 & 1.03 & 4.74 $\pm$ 2.69 & 0.00 \\
OPRK1 & 0.00 & 0.00 & 4.35 & 4.35 & 0.00 & 0.00 & 4.17 & 4.17 & 4.17 & 4.17 & 4.17 & 4.17 \\
PKM2 & 2.70 $\pm$ 2.06 & 0.37 & 3.83 $\pm$ 2.55 & 1.65 & 4.74 $\pm$ 3.58 & 2.39 & 3.03 $\pm$ 1.83 & 4.76 & 3.42 $\pm$ 1.95 & 5.49 & 4.01 $\pm$ 2.53 & 7.14 \\
PPARG & 2.49 $\pm$ 4.19 & 0.00 & 5.81 $\pm$ 6.03 & 7.74 & 6.64 $\pm$ 5.35 & 15.49 & 5.16 $\pm$ 4.98 & 8.15 & 7.33 $\pm$ 3.51 & 20.37 & 5.97 $\pm$ 5.52 & 12.22 \\
TP53 & 0.43 $\pm$ 0.67 & 0.00 & 2.16 $\pm$ 1.95 & 5.18 & 0.86 $\pm$ 1.34 & 2.59 & 0.25 $\pm$ 0.62 & 0.00 & 0.00 $\pm$ 0.00 & 0.00 & 0.00 $\pm$ 0.00 & 0.00 \\
VDR & 8.55 $\pm$ 0.08 & 8.49 & 0.34 $\pm$ 0.00 & 0.45 & 3.34 $\pm$ 0.08 & 3.17 & 2.07 & 2.07 & 0.69 & 0.69 & 0.69 & 0.69 \\
\midrule
Average & \textbf{2.77 $\pm$ 1.62} & 2.36 & 1.80 $\pm$ 1.89 & 2.88 & 2.25 $\pm$ 1.84 & \textbf{3.80} & \textbf{3.17 $\pm$ 2.09} & 3.48 & 2.11 $\pm$ 1.51 & \textbf{4.58} & 2.81 $\pm$ 2.03 & 4.28 \\
\bottomrule
\end{tabular}
}
\label{tab:roshambo2_geodiff_complete_cosine}
\end{table}

\begin{table}[ht]
\caption{\textbf{Virtual screening performance (EF1\%) obtained by different methods.} Each method reports per-reference Mean $\pm$ SD and Best-pooled EF1\%. ROSHAMBO2 score is based on the combination mode. GeoDiff uses 3D mode with cosine PED, and MoLFormer uses cosine PED. The best average values are shown in \textbf{bold}, and the second best are \underline{underlined}.}
\centering
\resizebox{\textwidth}{!}{
\begin{tabular}{l cc | cc | cc | cc | cc | cc }
\toprule
& \multicolumn{2}{c}{2D ECFP4 similarity} & \multicolumn{2}{c}{ROCS 3D similarity} & \multicolumn{2}{c}{Molecular docking} & \multicolumn{2}{c}{ROSHAMBO2 score} & \multicolumn{2}{c}{GeoDiff cosine distance} & \multicolumn{2}{c}{MoLFormer cosine distance} \\
\cmidrule(lr){2-3} \cmidrule(lr){4-5} \cmidrule(lr){6-7} \cmidrule(lr){8-9} \cmidrule(lr){10-11} \cmidrule(lr){12-13}
Target & Mean $\pm$ SD & Best-pooled & Mean $\pm$ SD & Best-pooled & Mean $\pm$ SD & Best-pooled & Mean $\pm$ SD & Best-pooled & Mean $\pm$ SD & Best-pooled & Mean $\pm$ SD & Best-pooled \\
\midrule
ADRB2 & 18.38 $\pm$ 6.62 & 17.65 & 9.76 $\pm$ 5.35 & 20.00 & 2.94 $\pm$ 3.14 & 5.88 & 3.91 $\pm$ 3.24 & 12.50 & 2.21 $\pm$ 3.04 & 5.88 & 16.17 $\pm$ 4.16 & 17.64 \\
ALDH1 & 1.40 $\pm$ 0.50 & 2.61 & 1.20 $\pm$ 0.39 & 1.85 & 1.19 $\pm$ 0.10 & 0.96 & 1.18 $\pm$ 0.52 & 0.68 & 1.20 $\pm$ 0.27 & 1.09 & 1.67 $\pm$ 0.32 & 1.95 \\
ESR1\_ago & 7.52 $\pm$ 6.09 & 7.69 & 4.61 $\pm$ 4.86 & 7.69 & 0.00 & 0.00 & 1.82 $\pm$ 3.61 & 8.47 & 3.01 $\pm$ 3.82 & 15.06 & 13.05 $\pm$ 8.76 & 7.53 \\
ESR1\_ant & 1.39 $\pm$ 1.06 & 0.98 & 1.66 $\pm$ 1.55 & 3.92 & 0.91 $\pm$ 0.78 & 0.98 & 2.53 $\pm$ 2.06 & 4.00 & 2.39 $\pm$ 1.85 & 5.57 & 1.11 $\pm$ 1.75 & 0.00 \\
FEN1 & 1.08 & 1.08 & 0.27 & 0.27 & 4.34 & 4.34 & 2.17 & 2.17 & 0.83 & 0.83 & 1.39 & 1.39 \\
GBA & 2.01 $\pm$ 1.41 & 3.61 & 0.72 $\pm$ 0.35 & 0.86 & 5.72 $\pm$ 1.97 & 8.43 & 1.58 $\pm$ 0.54 & 1.21 & 1.37 $\pm$ 0.30 & 3.05 & 4.27 $\pm$ 0.50 & 4.27 \\
IDH1 & 1.83 $\pm$ 3.09 & 5.13 & 0.86 $\pm$ 2.32 & 0.00 & 0.73 $\pm$ 1.20 & 0.00 & 1.01 $\pm$ 1.71 & 0.00 & 0.73 $\pm$ 1.20 & 2.56 & 1.65 $\pm$ 1.91 & 0.00 \\
KAT2A & 0.17 $\pm$ 0.30 & 0.52 & 1.48 $\pm$ 0.07 & 1.55 & 1.89 $\pm$ 0.59 & 2.58 & 1.55 $\pm$ 0.73 & 1.04 & 0.77 $\pm$ 0.36 & 1.03 & 2.83 $\pm$ 1.82 & 4.12 \\
MAPK1 & 1.20 $\pm$ 0.74 & 1.95 & 1.96 $\pm$ 1.01 & 5.19 & 1.32 $\pm$ 0.53 & 1.62 & 1.35 $\pm$ 0.67 & 2.28 & 1.95 $\pm$ 0.86 & 1.95 & 3.90 $\pm$ 1.98 & 4.55 \\
MTORC1 & 0.28 $\pm$ 0.48 & 0.00 & 1.38 $\pm$ 1.29 & 3.23 & 1.03 $\pm$ 0.65 & 1.03 & 1.04 $\pm$ 0.47 & 1.04 & 1.65 $\pm$ 0.92 & 1.03 & 0.82 $\pm$ 0.86 & 2.06 \\
OPRK1 & 12.50 & 12.50 & 0.00 & 0.00 & 4.17 & 4.17 & 0.00 & 0.00 & 4.17 & 4.17 & 4.17 & 4.17 \\
PKM2 & 3.30 $\pm$ 2.73 & 7.95 & 4.86 $\pm$ 4.30 & 6.22 & 0.57 $\pm$ 0.37 & 0.18 & 4.74 $\pm$ 3.58 & 2.39 & 3.42 $\pm$ 1.95 & 5.49 & 4.17 $\pm$ 3.85 & 10.25 \\
PPARG & 3.65 $\pm$ 5.14 & 7.41 & 8.79 $\pm$ 5.24 & 18.52 & 5.92 $\pm$ 5.20 & 7.41 & 6.64 $\pm$ 5.35 & 15.49 & 7.33 $\pm$ 3.51 & 20.37 & 5.98 $\pm$ 5.73 & 28.52 \\
TP53 & 1.06 $\pm$ 0.95 & 0.00 & 0.88 $\pm$ 1.36 & 2.53 & 0.00 & 0.00 & 0.86 $\pm$ 1.34 & 2.59 & 0.00 $\pm$ 0.00 & 0.00 & 1.02 $\pm$ 1.85 & 0.00 \\
VDR & 3.37 & 3.37 & 0.35 & 0.35 & 0.34 $\pm$ 0.48 & 0.00 & 3.34 $\pm$ 0.08 & 3.17 & 0.69 & 0.69 & 5.74 & 5.74 \\
\midrule
Average & \underline{3.94 $\pm$ 2.43} & \underline{4.83} & 2.59 $\pm$ 2.34 & 4.81 & 2.07 $\pm$ 1.36 & 2.51 & 2.25 $\pm$ 1.84 & 3.80 & 2.11 $\pm$ 1.51 & 4.58 & \textbf{4.53 $\pm$ 2.79} & \textbf{6.15} \\
\bottomrule
\end{tabular}
}
\label{tab:vs_comparison_mean_pooled_roshambo2_3D_cosine}
\end{table}


\begin{table}[ht]
\caption{\textbf{Complete virtual screening performance (EF1\%) obtained by different modes of ROSHAMBO2 similarity and GeoDiff Euclidean PED.} The best average mean and best-pooled EF1\% values for each approach are shown in \textbf{bold}.}
\centering
\resizebox{\textwidth}{!}{
\begin{tabular}{l cc | cc | cc | cc | cc | cc}
\toprule
& \multicolumn{2}{c}{ROSHAMBO2 color} & \multicolumn{2}{c}{ROSHAMBO2 shape} & \multicolumn{2}{c}{ROSHAMBO2 combination} & \multicolumn{2}{c}{GeoDiff 2D} & \multicolumn{2}{c}{GeoDiff 3D} & \multicolumn{2}{c}{GeoDiff concat} \\
\cmidrule(lr){2-3} \cmidrule(lr){4-5} \cmidrule(lr){6-7} \cmidrule(lr){8-9} \cmidrule(lr){10-11} \cmidrule(lr){12-13}
Target & Mean $\pm$ SD & Best-pooled & Mean $\pm$ SD & Best-pooled & Mean $\pm$ SD & Best-pooled & Mean $\pm$ SD & Best-pooled & Mean $\pm$ SD & Best-pooled & Mean $\pm$ SD & Best-pooled \\
\midrule
ADRB2 & 10.94 $\pm$ 5.54 & 12.50 & 0.78 $\pm$ 2.21 & 6.25 & 3.91 $\pm$ 3.24 & 12.50 & 14.70 $\pm$ 4.45 & 11.76 & 2.21 $\pm$ 3.04 & 5.88 & 6.62 $\pm$ 3.77 & 11.76 \\
ALDH1 & 0.98 $\pm$ 0.04 & 1.02 & 1.23 $\pm$ 0.30 & 0.88 & 1.18 $\pm$ 0.52 & 0.68 & 0.92 $\pm$ 0.17 & 0.92 & 1.23 $\pm$ 0.29 & 1.09 & 1.08 $\pm$ 0.25 & 1.29 \\
ESR1\_ago & 3.03 $\pm$ 4.21 & 0.00 & 1.82 $\pm$ 4.91 & 8.47 & 1.82 $\pm$ 3.61 & 8.47 & 4.02 $\pm$ 6.28 & 0.00 & 3.01 $\pm$ 3.82 & 22.59 & 4.02 $\pm$ 4.82 & 15.06 \\
ESR1\_ant & 0.73 $\pm$ 0.88 & 1.00 & 1.67 $\pm$ 1.34 & 1.00 & 2.53 $\pm$ 2.06 & 4.00 & 1.19 $\pm$ 1.02 & 2.23 & 2.31 $\pm$ 1.88 & 5.57 & 2.71 $\pm$ 1.56 & 2.23 \\
FEN1 & 5.16 & 5.16 & 0.54 & 0.54 & 2.17 & 2.17 & 2.78 & 2.78 & 0.83 & 0.83 & 2.50 & 2.50 \\
GBA & 2.30 $\pm$ 0.66 & 3.64 & 0.36 $\pm$ 0.54 & 1.21 & 1.58 $\pm$ 0.54 & 1.21 & 3.20 $\pm$ 1.35 & 3.05 & 1.37 $\pm$ 0.30 & 3.05 & 2.13 $\pm$ 0.35 & 1.83 \\
IDH1 & 0.20 $\pm$ 0.73 & 0.00 & 1.01 $\pm$ 1.71 & 0.00 & 1.01 $\pm$ 1.71 & 0.00 & 2.20 $\pm$ 2.22 & 7.69 & 0.73 $\pm$ 1.20 & 2.56 & 1.28 $\pm$ 1.67 & 2.56 \\
KAT2A & 1.04 $\pm$ 0.73 & 1.55 & 1.30 $\pm$ 1.10 & 2.07 & 1.55 $\pm$ 0.73 & 1.04 & 0.00 $\pm$ 0.00 & 0.00 & 1.03 $\pm$ 0.73 & 1.55 & 0.26 $\pm$ 0.36 & 0.52 \\
MAPK1 & 0.60 $\pm$ 0.48 & 0.65 & 1.56 $\pm$ 1.31 & 1.30 & 1.35 $\pm$ 0.67 & 2.28 & 1.42 $\pm$ 1.26 & 1.30 & 1.92 $\pm$ 0.83 & 1.95 & 1.95 $\pm$ 0.89 & 2.27 \\
MTORC1 & 2.37 $\pm$ 0.82 & 1.04 & 0.28 $\pm$ 0.67 & 2.09 & 1.04 $\pm$ 0.47 & 1.04 & 2.47 $\pm$ 0.92 & 2.06 & 1.65 $\pm$ 0.92 & 1.03 & 4.74 $\pm$ 2.69 & 0.00 \\
OPRK1 & 0.00 & 0.00 & 4.35 & 4.35 & 0.00 & 0.00 & 4.17 & 4.17 & 8.33 & 8.33 & 4.17 & 4.17 \\
PKM2 & 2.70 $\pm$ 2.06 & 0.37 & 3.83 $\pm$ 2.55 & 1.65 & 4.74 $\pm$ 3.58 & 2.39 & 3.05 $\pm$ 1.79 & 5.13 & 3.58 $\pm$ 2.06 & 6.59 & 4.01 $\pm$ 2.53 & 7.14 \\
PPARG & 2.49 $\pm$ 4.19 & 0.00 & 5.81 $\pm$ 6.03 & 7.74 & 6.64 $\pm$ 5.35 & 15.49 & 4.89 $\pm$ 5.15 & 8.15 & 6.52 $\pm$ 3.71 & 20.37 & 5.97 $\pm$ 5.52 & 12.22 \\
TP53 & 0.43 $\pm$ 0.67 & 0.00 & 2.16 $\pm$ 1.95 & 5.18 & 0.86 $\pm$ 1.34 & 2.59 & 0.25 $\pm$ 0.62 & 0.00 & 0.00 $\pm$ 0.00 & 0.00 & 0.00 $\pm$ 0.00 & 0.00 \\
VDR & 8.55 $\pm$ 0.08 & 8.49 & 0.34 $\pm$ 0.00 & 0.45 & 3.34 $\pm$ 0.08 & 3.17 & 2.07 & 2.07 & 0.69 & 0.69 & 0.69 & 0.69 \\
\midrule
Average & \textbf{2.77 $\pm$ 1.62} & 2.36 & 1.80 $\pm$ 1.89 & 2.88 & 2.25 $\pm$ 1.84 & \textbf{3.80} & \textbf{3.16 $\pm$ 2.10} & 3.42 & 2.36 $\pm$ 1.57 & \textbf{5.47} & 2.81 $\pm$ 2.03 & 4.28 \\
\bottomrule
\end{tabular}
}
\label{tab:roshambo2_geodiff_complete_Euclidean}
\end{table}

\begin{table}[ht]
\caption{\textbf{Virtual screening performance (EF1\%) obtained by different methods.} Each method reports per-reference Mean $\pm$ SD and Best-pooled EF1\%. ROSHAMBO2 score is based on the combination mode. GeoDiff uses 3D mode with Euclidean PED, and MoLFormer uses Euclidean PED. The best average values are shown in \textbf{bold}, and the second best are \underline{underlined}.}
\centering
\resizebox{\textwidth}{!}{
\begin{tabular}{l cc | cc | cc | cc | cc | cc }
\toprule
& \multicolumn{2}{c}{2D ECFP4 similarity} & \multicolumn{2}{c}{ROCS 3D similarity} & \multicolumn{2}{c}{Molecular docking} & \multicolumn{2}{c}{ROSHAMBO2 score} & \multicolumn{2}{c}{GeoDiff Euclidean distance} & \multicolumn{2}{c}{MoLFormer Euclidean distance} \\
\cmidrule(lr){2-3} \cmidrule(lr){4-5} \cmidrule(lr){6-7} \cmidrule(lr){8-9} \cmidrule(lr){10-11} \cmidrule(lr){12-13}
Target & Mean $\pm$ SD & Best-pooled & Mean $\pm$ SD & Best-pooled & Mean $\pm$ SD & Best-pooled & Mean $\pm$ SD & Best-pooled & Mean $\pm$ SD & Best-pooled & Mean $\pm$ SD & Best-pooled \\
\midrule
ADRB2 & 18.38 $\pm$ 6.62 & 17.65 & 9.76 $\pm$ 5.35 & 20.00 & 2.94 $\pm$ 3.14 & 5.88 & 3.91 $\pm$ 3.24 & 12.50 & 2.21 $\pm$ 3.04 & 5.88 & 11.76 $\pm$ 6.29 & 17.64 \\
ALDH1 & 1.40 $\pm$ 0.50 & 2.61 & 1.20 $\pm$ 0.39 & 1.85 & 1.19 $\pm$ 0.10 & 0.96 & 1.18 $\pm$ 0.52 & 0.68 & 1.23 $\pm$ 0.29 & 1.09 & 1.69 $\pm$ 0.29 & 1.99 \\
ESR1\_ago & 7.52 $\pm$ 6.09 & 7.69 & 4.61 $\pm$ 4.86 & 7.69 & 0.00 & 0.00 & 1.82 $\pm$ 3.61 & 8.47 & 3.01 $\pm$ 3.82 & 22.59 & 13.05 $\pm$ 9.64 & 0.00 \\
ESR1\_ant & 1.39 $\pm$ 1.06 & 0.98 & 1.66 $\pm$ 1.55 & 3.92 & 0.91 $\pm$ 0.78 & 0.98 & 2.53 $\pm$ 2.06 & 4.00 & 2.31 $\pm$ 1.88 & 5.57 & 0.96 $\pm$ 1.06 & 0.00 \\
FEN1 & 1.08 & 1.08 & 0.27 & 0.27 & 4.34 & 4.34 & 2.17 & 2.17 & 0.83 & 0.83 & 1.11 & 1.11 \\
GBA & 2.01 $\pm$ 1.41 & 3.61 & 0.72 $\pm$ 0.35 & 0.86 & 5.72 $\pm$ 1.97 & 8.43 & 1.58 $\pm$ 0.54 & 1.21 & 1.37 $\pm$ 0.30 & 3.05 & 4.11 $\pm$ 0.58 & 3.66 \\
IDH1 & 1.83 $\pm$ 3.09 & 5.13 & 0.86 $\pm$ 2.32 & 0.00 & 0.73 $\pm$ 1.20 & 0.00 & 1.01 $\pm$ 1.71 & 0.00 & 0.73 $\pm$ 1.20 & 2.56 & 2.20 $\pm$ 2.82 & 0.00 \\
KAT2A & 0.17 $\pm$ 0.30 & 0.52 & 1.48 $\pm$ 0.07 & 1.55 & 1.89 $\pm$ 0.59 & 2.58 & 1.55 $\pm$ 0.73 & 1.04 & 1.03 $\pm$ 0.73 & 1.55 & 2.32 $\pm$ 0.36 & 2.58 \\
MAPK1 & 1.20 $\pm$ 0.74 & 1.95 & 1.96 $\pm$ 1.01 & 5.19 & 1.32 $\pm$ 0.53 & 1.62 & 1.35 $\pm$ 0.67 & 2.28 & 1.92 $\pm$ 0.83 & 1.95 & 3.69 $\pm$ 2.23 & 4.22 \\
MTORC1 & 0.28 $\pm$ 0.48 & 0.00 & 1.38 $\pm$ 1.29 & 3.23 & 1.03 $\pm$ 0.65 & 1.03 & 1.04 $\pm$ 0.47 & 1.04 & 1.65 $\pm$ 0.92 & 1.03 & 2.27 $\pm$ 0.86 & 1.03 \\
OPRK1 & 12.50 & 12.50 & 0.00 & 0.00 & 4.17 & 4.17 & 0.00 & 0.00 & 8.33 & 8.33 & 4.17 & 4.17 \\
PKM2 & 3.30 $\pm$ 2.73 & 7.95 & 4.86 $\pm$ 4.30 & 6.22 & 0.57 $\pm$ 0.37 & 0.18 & 4.74 $\pm$ 3.58 & 2.39 & 3.58 $\pm$ 2.06 & 6.59 & 4.39 $\pm$ 3.88 & 10.44 \\
PPARG & 3.65 $\pm$ 5.14 & 7.41 & 8.79 $\pm$ 5.24 & 18.52 & 5.92 $\pm$ 5.20 & 7.41 & 6.64 $\pm$ 5.35 & 15.49 & 6.52 $\pm$ 3.71 & 20.37 & 6.52 $\pm$ 7.35 & 24.45 \\
TP53 & 1.06 $\pm$ 0.95 & 0.00 & 0.88 $\pm$ 1.36 & 2.53 & 0.00 & 0.00 & 0.86 $\pm$ 1.34 & 2.59 & 0.00 $\pm$ 0.00 & 0.00 & 1.53 $\pm$ 1.93 & 0.00 \\
VDR & 3.37 & 3.37 & 0.35 & 0.35 & 0.34 $\pm$ 0.48 & 0.00 & 3.34 $\pm$ 0.08 & 3.17 & 0.69 & 0.69 & 4.94 & 4.94 \\
\midrule
Average & \textbf{3.94 $\pm$ 2.43} & 4.83 & 2.59 $\pm$ 2.34 & 4.81 & 2.07 $\pm$ 1.36 & 2.51 & 2.25 $\pm$ 1.84 & 3.80 & 2.36 $\pm$ 1.57 & \textbf{5.47} & \textbf{4.31 $\pm$ 3.11} & \underline{5.08} \\
\bottomrule
\end{tabular}
}
\label{tab:vs_comparison_mean_pooled_roshambo2_3D_Euclidean}
\end{table}

\clearpage
\section{Training Dynamics of Generative Frameworks with different Rewards}

\begin{figure}[H]
    \centering
    \includegraphics[width=.9\linewidth]{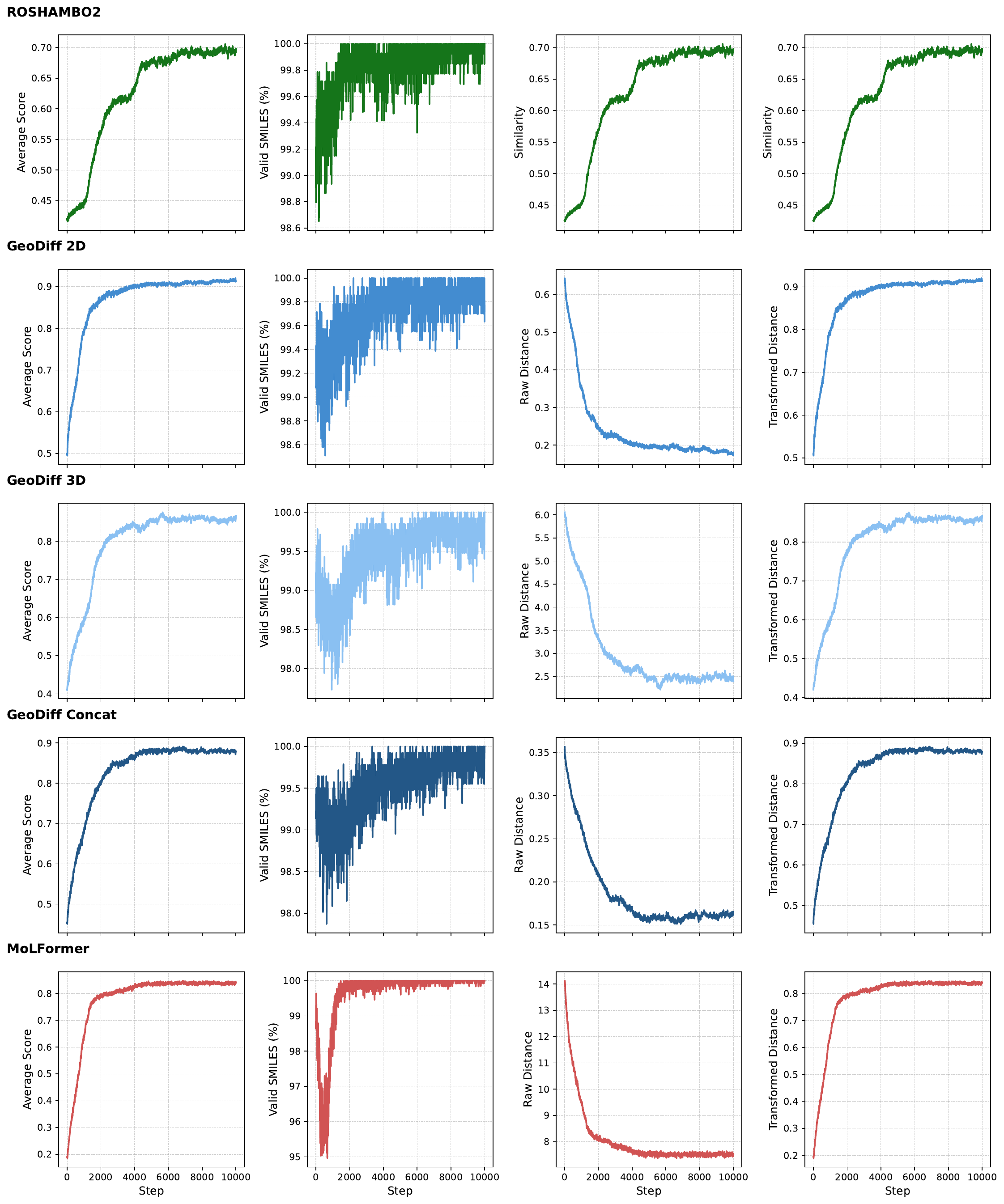}
    \caption{\textbf{Training dynamics of REINVENT with different reward components.} Average score, valid SMILES rate, and similarity or distance metrics are tracked during training for ROSHAMBO2, GeoDiff PED variants, and MoLFormer PED. Curves are computed by averaging scores within each training step and applying a moving average smoothing to reduce stochastic variation from sampling.}
    \label{fig:reinvent_training}
\end{figure}

\begin{figure}[H]
    \centering
    \includegraphics[width=.6\linewidth]{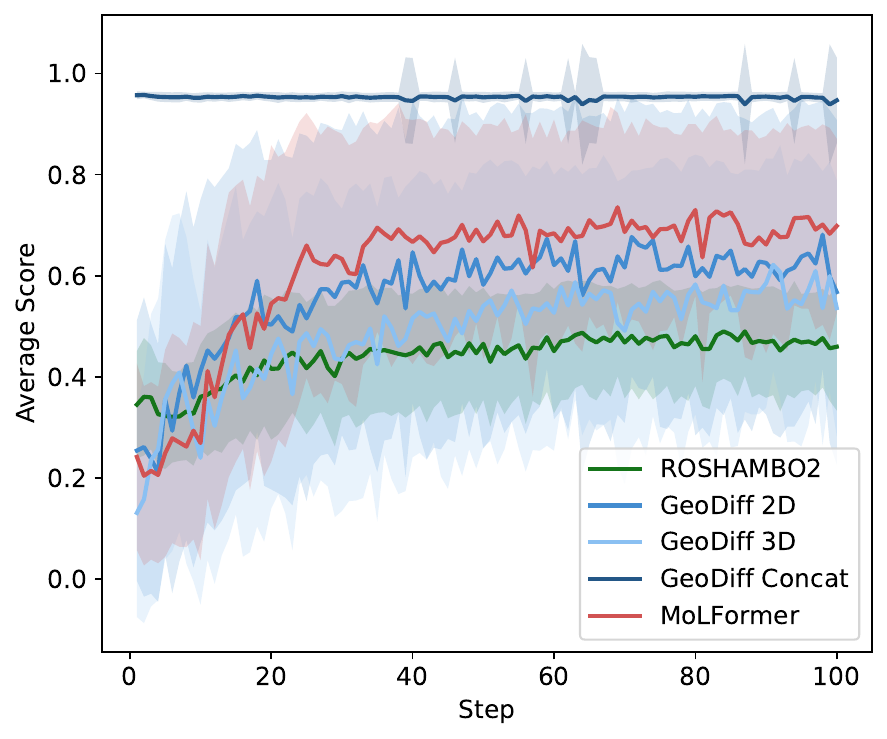}
    \caption{\textbf{Training dynamics of SynFormer with different reward components.} Curves are computed by averaging scores within each training step.}
    \label{fig:synformer_training}
\end{figure}

\section{Statistics of Generated Molecules}
\label{app:statistics_mols}

As mentioned in Appendix \ref{app:setting}, in each run, REINVENT generates 1,280,000 molecules and SynFormer generates 12,800 molecules. Table \ref{tab:generate_statistics} shows all the unique molecules and scaffolds generated by REINVENT and SynFormer.

\begin{table}[ht]
    \centering
    \caption{REINVENT and SynFormer generated unique molecules, unique ring-linker scaffolds, and the unique scaffold ratio.}
    \begin{tabular}{llcccc}
    \toprule
         Model & Method & Unique molecules & Unique scaffold & Unique scaffold ratio ($\uparrow$) & Running time (h) ($\downarrow$)\\
    \midrule
    \multirow{6}{*}{REINVENT}
        & ROSHAMBO2       & 277,116 & 100,983 & 36.44\% & 47 \\
        & GeoDiff 2D      & 211,835 & 84,764 & 40.01\% & 31 \\
        & GeoDiff 3D      & 421,879 & 230,346 & \textbf{54.60\%} & 31 \\
        & GeoDiff Concat  & 338,333 & 119,441 & 35.30\% & 31 \\
        & MoLFormer       & 179,598 & 86,091 & 47.94\% & \textbf{14} \\
    \midrule
    \multirow{6}{*}{SynFormer}
        & ROSHAMBO2       & 6,813 & 1,459 & 21.41\% & 1.33 \\
        & GeoDiff 2D      & 8,853 & 3,827 & \textbf{43.23\%} & 0.70 \\
        & GeoDiff 3D      & 7,999 & 1,532 & 19.15\% & 0.70 \\
        & GeoDiff Concat  & 6,089 & 1,081 & 17.75\% & 0.70 \\
        & MoLFormer       & 7,357 & 1,003 & 13.63\% & \textbf{0.65} \\
    \bottomrule
    \end{tabular}
    \label{tab:generate_statistics}
\end{table}

\section{Drug-likeness of Top-5,000 Generated Molecules}
\label{app:drug_likeness}

We followed the evaluation protocol of Wang et al. \yrcite{molecules28114430} to assess the physicochemical properties of the generated compounds. Specifically, for the top 5,000 generated molecules, we computed five physicochemical properties, including molecular weight (MW), topological polar surface area (TPSA), partition coefficient (LogP, indicating hydrophobicity), number of hydrogen bond donors (HBD), and number of hydrogen bond acceptors (HBA), along with quantitative estimate of drug-likeness (QED) and synthetic accessibility (SA) scores. The minimum and maximum values of these physicochemical properties are reported in Table \ref{tab:physchem_top5000}, showing that the generated molecules cover a broad range of physicochemical space. In addition, we evaluated the proportion of molecules that fall within desirable drug-like ranges for each property (Table \ref{tab:druglikeness_percent}).

\begin{table}[ht]
\centering
\caption{Physical-chemical properties of top-5,000 generated molecules by REINVENT and SynFormer.}
\label{tab:physchem_top5000}
\resizebox{\textwidth}{!}{
\begin{tabular}{llcccccccccc}
\toprule
Model & Method & MW$_{min}$ & MW$_{max}$ & TPSA$_{min}$ & TPSA$_{max}$ & LogP$_{min}$ & LogP$_{max}$ & HBD$_{min}$ & HBD$_{max}$ & HBA$_{min}$ & HBA$_{max}$ \\
\midrule

\multirow{5}{*}{REINVENT}
& ROSHAMBO2 & 245.12 & 469.04 & 41.57 & 136.82 & -0.26 & 5.31 & 0.00 & 4.00 & 2.00 & 7.00 \\
& GeoDiff 2D & 258.15 & 478.24 & 45.48 & 134.31 & 0.96 & 4.56 & 1.00 & 5.00 & 1.00 & 5.00 \\
& GeoDiff 3D & 325.19 & 466.11 & 38.13 & 165.65 & 0.09 & 5.23 & 0.00 & 5.00 & 2.00 & 8.00 \\
& GeoDiff Concat & 301.18 & 433.19 & 54.34 & 147.20 & 0.39 & 4.17 & 1.00 & 4.00 & 2.00 & 5.00 \\
& MoLFormer & 307.15 & 496.25 & 39.34 & 151.55 & 0.11 & 5.14 & 0.00 & 5.00 & 2.00 & 6.00 \\

\midrule
\multirow{5}{*}{SynFormer}
& ROSHAMBO2 & 128.04 & 650.02 & 0.00 & 252.98 & -1.79 & 8.28 & 0.00 & 7.00 & 0.00 & 10.00 \\
& GeoDiff 2D & 261.18 & 1116.36 & 31.06 & 278.13 & -0.57 & 12.75 & 0.00 & 8.00 & 1.00 & 14.00 \\
& GeoDiff 3D & 224.15 & 690.99 & 6.48 & 193.57 & -0.68 & 8.07 & 0.00 & 6.00 & 1.00 & 9.00 \\
& GeoDiff Concat & 73.09 & 1089.41 & 0.00 & 306.02 & -3.83 & 11.91 & 0.00 & 6.00 & 0.00 & 18.00 \\
& MoLFormer & 219.08 & 823.37 & 31.06 & 227.01 & -0.05 & 10.77 & 0.00 & 6.00 & 1.00 & 9.00 \\
\bottomrule
\end{tabular}
}
\end{table}

\begin{table}[ht]
\centering
\caption{Percentages of top-5,000 generated molecules by REINVENT and SynFormer in good drug-likeness range.}
\label{tab:druglikeness_percent}
\resizebox{\textwidth}{!}{
\begin{tabular}{llccccccc}
\toprule
Model & Method & MW ([200, 500]) & TPSA ([20, 130]) & LogP ([$-1$, 6]) & HBD ([0, 5]) & HBA ([0, 10]) & QED ([0.4, 1]) & SA ([1, 5])  \\

\midrule
\multirow{5}{*}{REINVENT}
& ROSHAMBO2 & 100.00\% & 99.82\% & 100.00\% & 100.00\% & 100.00\% & 100.00\% & 100.00\%  \\
& GeoDiff 2D & 100.00\% & 99.98\% & 100.00\% & 100.00\% & 100.00\% & 99.98\% & 100.00\%  \\
& GeoDiff 3D & 100.00\% & 94.38\% & 100.00\% & 100.00\% & 100.00\% & 99.98\% & 100.00\% \\
& GeoDiff Concat & 100.00\% & 97.56\% & 100.00\% & 100.00\% & 100.00\% & 100.00\% & 100.00\%  \\
& MoLFormer & 100.00\% & 98.40\% & 100.00\% & 100.00\% & 100.00\% & 99.92\% & 100.00\%  \\

\midrule
\multirow{5}{*}{SynFormer}
& ROSHAMBO2 & 92.06\% & 94.18\% & 99.26\% & 99.82\% & 100.00\% & 94.24\% & 100.00\%  \\
& GeoDiff 2D & 57.18\% & 84.92\% & 75.74\% & 98.18\% & 99.00\% & 49.18\% & 98.92\%  \\
& GeoDiff 3D & 97.04\% & 98.06\% & 99.02\% & 99.92\% & 100.00\% & 86.42\% & 100.00\%  \\
& GeoDiff Concat & 36.10\% & 84.26\% & 96.40\% & 99.82\% & 99.62\% & 92.84\% & 99.72\%  \\
& MoLFormer & 94.48\% & 98.36\% & 95.78\% & 99.92\% & 100.00\% & 75.16\% & 100.00\%  \\

\bottomrule
\end{tabular}
}
\end{table}

\clearpage
\section{Top-5 Generated Molecules by REINVENT and SynFormer}
\label{app:top5}

\begin{figure}[H]
    \centering
    \includegraphics[width=1.\linewidth]{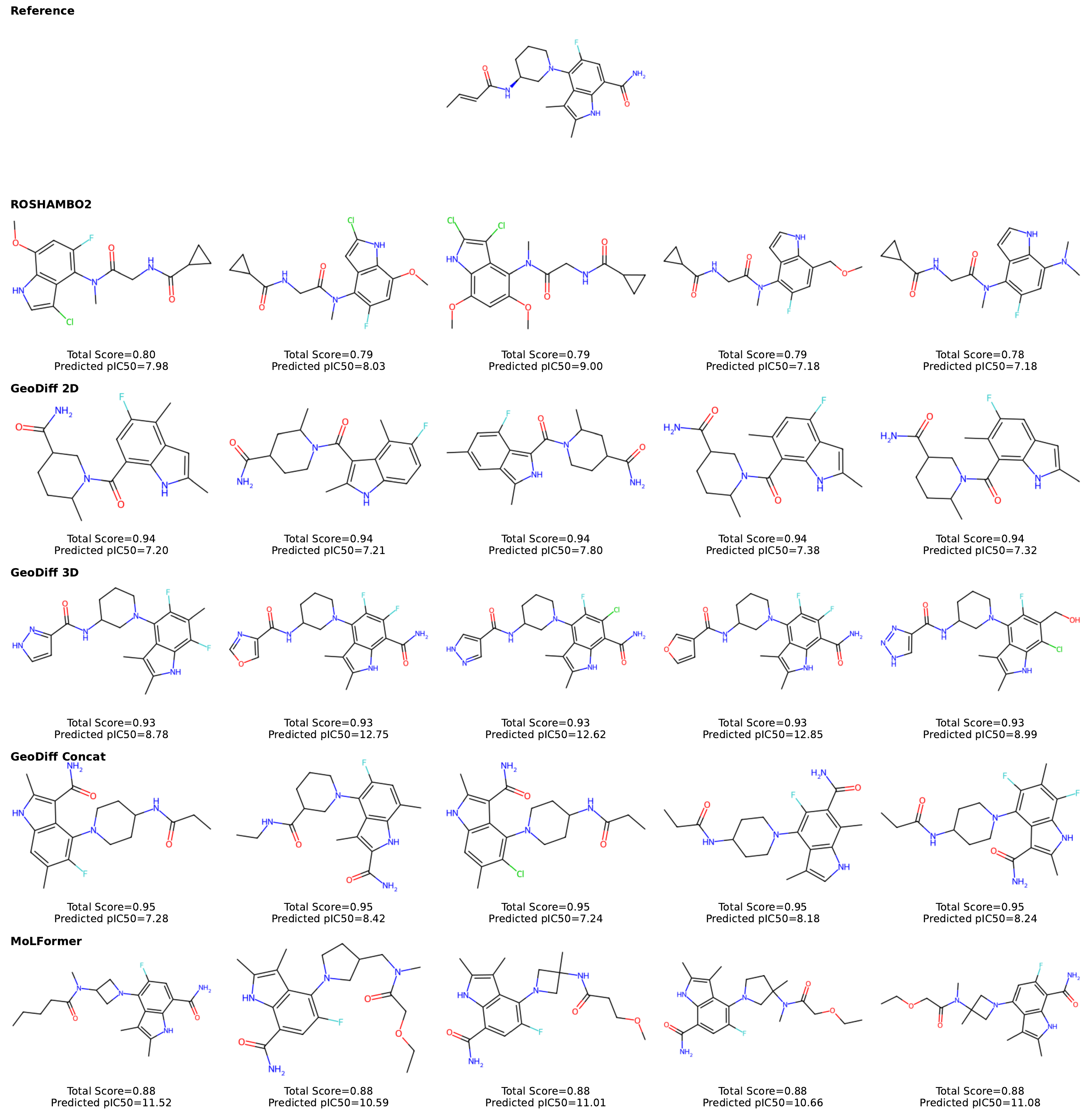}
    \caption{The reference molecule and REINVENT-generated molecules in top-5 total score across methods.}
    \label{fig:top5_reinvent}
\end{figure}

\begin{figure}[H]
    \centering
    \includegraphics[width=1.\linewidth]{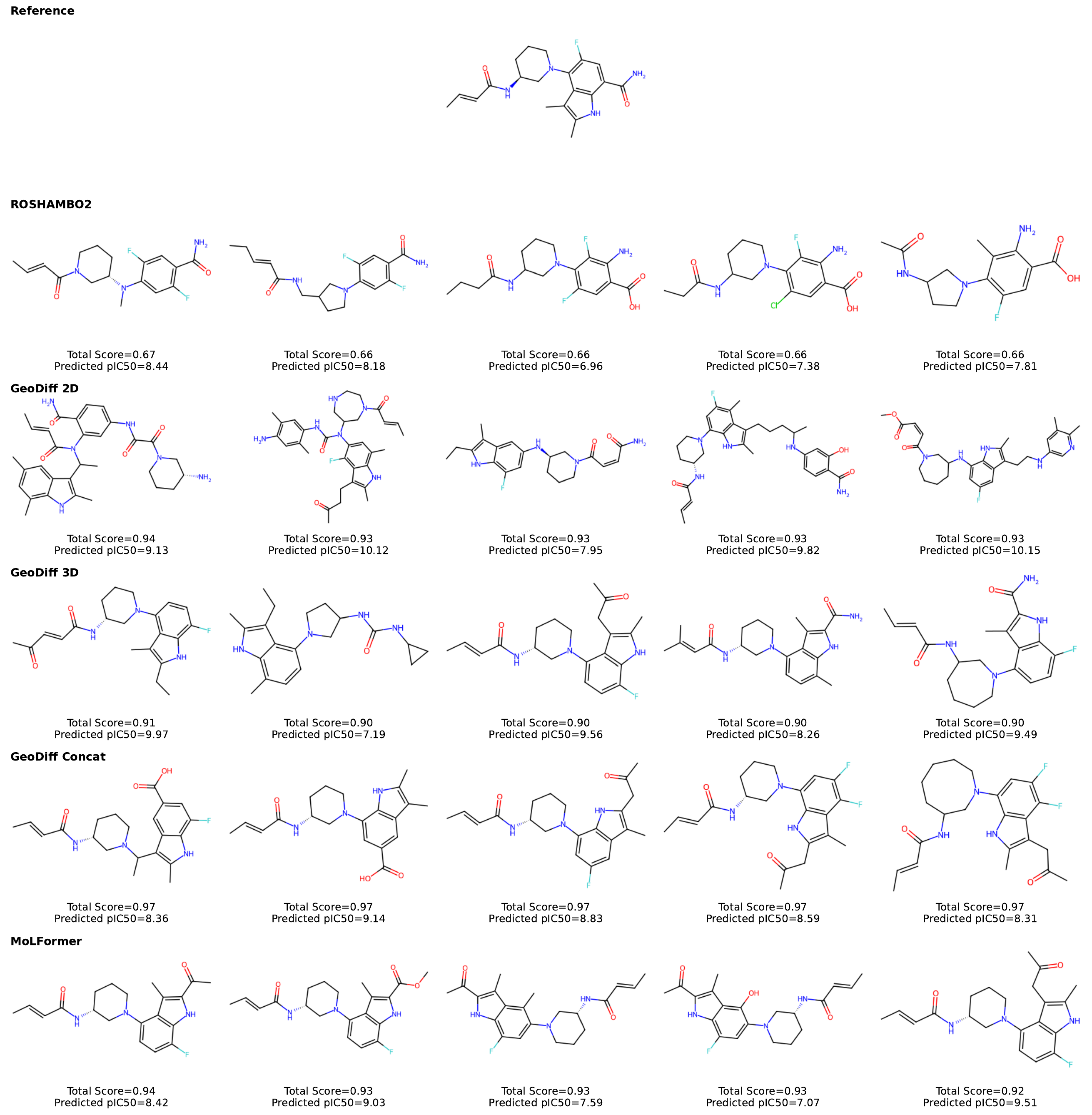}
    \caption{The reference molecule and SynFormer-generated molecules in top-5 total score across methods.}
    \label{fig:top5_synformer}
\end{figure}

\end{document}